%% file: example_paper.tex
\documentclass[letterpaper,journal]{IEEEtran}
\usepackage{amsmath}
\usepackage{amssymb}
\usepackage{mathtools}
\usepackage{amsthm}

\usepackage{xcolor}
\usepackage{xspace}
\usepackage{url}
\usepackage{multirow}
\usepackage{adjustbox}
\usepackage{makecell}
\usepackage{colortbl}

\usepackage{array}
\usepackage[capitalize,noabbrev]{cleveref}
\usepackage{wrapfig}
\usepackage{caption}

\usepackage{threeparttable}
\usepackage{rotating}
\usepackage{lipsum}
\usepackage{tikz}
\usetikzlibrary{tikzmark}
\usepackage{enumitem}
\usepackage{microtype}
\usepackage{caption}
\usepackage{algorithm,algorithmic}

\usepackage{graphicx}
\usepackage{booktabs}
\usepackage{subcaption}
\usepackage[textsize=tiny]{todonotes}

\theoremstyle{plain}

\theoremstyle{definition}

\theoremstyle{remark}

\def\model{LDM4TS\xspace}

\newcommand{\boldres}[1]{{\textbf{\textcolor{red}{#1}}}}
\newcommand{\secondres}[1]{{\underline{\textcolor{blue}{#1}}}}

\newcommand{\citeyearpar}[1]{(\the\year)}

\begin{document}

\title{Vision-Enhanced Time Series Forecasting via Latent Diffusion Models}

\author{Weilin Ruan, Siru Zhong, Haomin Wen and~Yuxuan~Liang
\thanks{W. Ruan, S. Zhong, and Y. Liang are with the Hong Kong University of Science and Technology (Guangzhou), China.
Corresponding author: Y. Liang (yuxliang@outlook.com)}
\thanks{H. Wen is with Carnegie Mellon University, USA.}
}

\maketitle

\input{contents/00abstract}
\input{contents/01introduction}
\input{contents/02relatedwork}
\input{contents/03method}
\input{contents/04experiments}
\input{contents/05conclusion}

\bibliographystyle{IEEEtran}
\bibliography{example_paper}

\nocite{shen2023non,shen2024multi,suhtimeautodiff,yuvision,bilovs2022modeling,
zhang2023mixed,suh2023autodiff,zhou2024sdformer,ruan2024cross,ruan2024low,
chang2023llm4ts,jia2024gpt4mts,yuan2024diffusion,liu2024unitime,liu2024time,
wang2024timemixer++,zhang2024self, jin2024position, hu2024attractor,cai2024msgnet}

\input{contents/06appendix}
\end{document}

%% file: contents/00abstract.tex
\begin{abstract}
Diffusion models have recently emerged as powerful frameworks for generating high-quality images. While recent studies have explored their application to time series forecasting, these approaches face significant challenges in cross-modal modeling and transforming visual information effectively to capture temporal patterns. In this paper, we propose \model, a novel framework that leverages the powerful image reconstruction capabilities of latent diffusion models for vision-enhanced time series forecasting. Instead of introducing external visual data, we are the first to use complementary transformation techniques to convert time series into multi-view visual representations, allowing the model to exploit the rich feature extraction capabilities of the pre-trained vision encoder. Subsequently, these representations are reconstructed using a latent diffusion model with a cross-modal conditioning mechanism as well as a fusion module. Experimental results demonstrate that \model outperforms various specialized forecasting models for time series forecasting tasks.
\end{abstract}

%% file: contents/01introduction.tex
\vspace{-2em}
\section{Introduction}
Time Series Forecasting (TSF) is a critical task in numerous real-world applications~\cite{jin2024survey}, such as demand planning~\cite{leonard2001promotional}, energy load estimation~\cite{liu2023sadi}, climate modeling~\cite{schneider1974climate}, and traffic flow management~\cite{zheng2006short}. Deep learning models have significantly advanced time series forecasting by capturing intricate temporal dependencies. Early methods introduced sequential modeling capabilities~\cite{cho2014learning,hochreiter1997long}, while later architectures such as Transformers have improved the modeling of long-range dependencies and efficiency~\cite{nie2022time,zhou2021informer,zhou2022fedformer,wu2021autoformer,woo2022etsformer}. Despite their success, these models lack the ability to model the underlying uncertainty behind the time series.

In parallel with these developments, diffusion models have emerged as powerful generative frameworks, excelling in tasks like text-to-image generation~\cite{rombach2022high}, image synthesis~\cite{ho2020denoising}, and super-resolution~\cite{saharia2022palette}. Their iterative denoising process demonstrates exceptional capability in modeling complex distributions, prompting recent applications to time series modeling~\cite{rasul2021autoregressive,shen2024multi,shen2023non,tashiro2021csdi,yan2021scoregrad}. However, these models face significant challenges in time series forecasting as they are inherently designed for spatially structured data like images, while time series exhibits fundamentally different characteristics such as sequential dependencies and non-stationarity~\cite{box2015time,hamilton2020time}. The potential of transforming temporal patterns into structured visual representations, rather than directly processing raw sequences, remains largely unexplored for leveraging diffusion-based temporal dynamics.

\begin{figure}[!t]
  \centering
  \includegraphics[width=\linewidth]{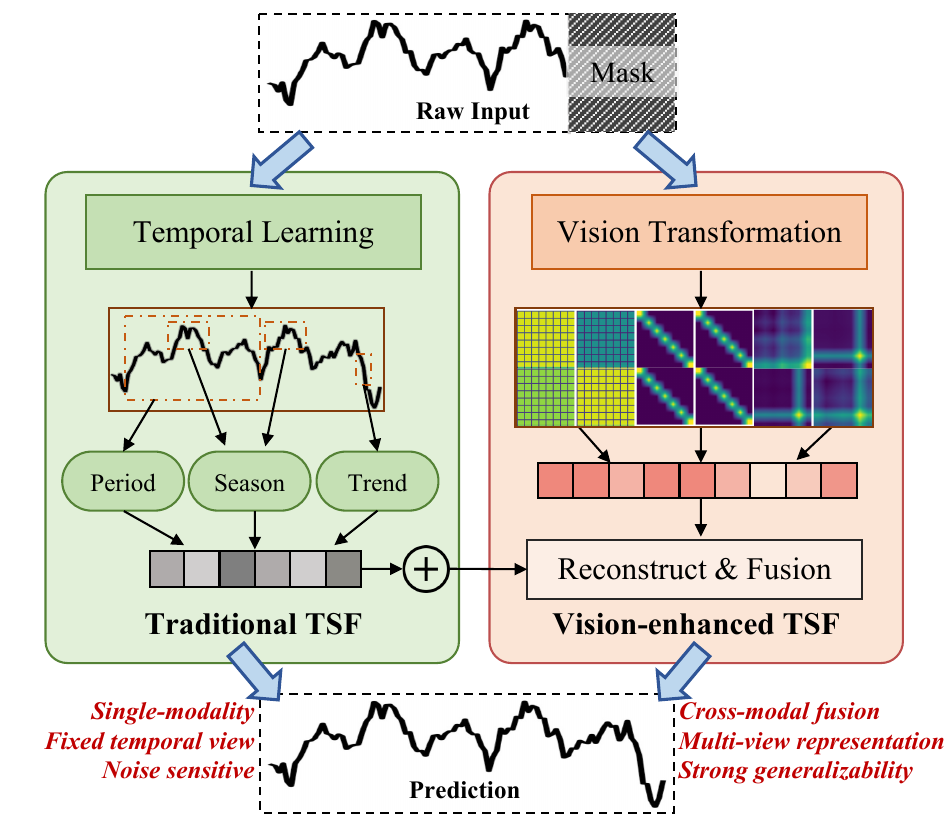}
  \caption{Comparison between traditional TSF methods and vision-enhanced approach, highlighting our method leverages multi-view visual representations to enhance TSF.}
\vspace{-1.5em}
\label{fig:method_comparison}
\end{figure}
Meanwhile, previous research has demonstrated that transforming time series data into visual representations can effectively preserve crucial temporal structures, including periodicity, seasonality, and trends~\cite{eckmann1995recurrence, Oord2016WaveNetAG,wang2015encoding}. These visual representations enable the application of pre-trained vision models, which are adept at capturing local and global patterns, to time series forecasting~\cite{wu2022timesnet,chen2024visionts}. However, current time series visual modeling approaches still face several critical limitations: (i) Most existing methods rely heavily on single-view transformations, which may not fully capture the complex temporal dynamics inherent in time series data; (ii) While visual features extracted from time series data contain rich structural information, they lack specific mechanisms for temporal reasoning and future state prediction; (iii) The integration of uncertainty quantification and probabilistic forecasting remains largely unexplored in vision-based approaches. 

These observations highlight the need for a unified framework that synergistically integrates the complementary strengths of visual representations, diffusion models, and temporal modeling. To bridge this gap, we present \model, the first attempt to leverage vision-enhanced time series encoding into the latent diffusion model, by reformulating time series forecasting as an image reconstruction and cross-modal fusion task as illustrated in Figure~\ref{fig:method_comparison}. \model combines the probabilistic uncertainty modeling capabilities with sophisticated multi-view vision-enhanced temporal dependency learning. Through multiple complementary vision transformations and a specialized temporal fusion mechanism, our approach effectively captures both global patterns and local dynamics while maintaining efficiency.

Specifically, we first transform raw time series data into multi-view visual representations using structured encodings, including segmentation, recurrence plots and Gramian angular fields. These visual representations are then mapped into a low-dimensional latent space, where a pre-trained latent diffusion model progressively denoises the latent variables. To further increase the flexibility of the model, we condition the frequency embedding and textual embedding to capture domain-specific knowledge or statistical properties of the time series through cross-attention mechanisms. Finally, a temporal projection and fusion module is introduced to extract temporal dependencies from the reconstructed representations and predict future time series. In summary, the key contributions of this work are as follows:

\noindent \textbf{1) Multi-view Visual Representations:} 
We present the first work that utilizes complementary strategies to transform temporal features into multi-view visual representations, enabling both crucial temporal property preservation and enhanced complex temporal pattern recognition via pre-trained vision models.

\noindent \textbf{2) Latent Diffusion Framework with Multi-modal Conditions:} We pioneer \model, a unified framework that leverages latent diffusion processes, combined with visual representations and cross-modal conditioning mechanism, to effectively integrate visual pattern extraction, uncertainty modeling, and temporal dependency learning for TSF.

\noindent \textbf{3) Comprehensive Empirical Validation:} Extensive experiments verify that \model achieves state-of-the-art performance on diverse datasets, outperforming existing diffusion-based methods by at least 65.2\% in terms of MSE while achieving at least 15.7\% improvement over the runner-up.

%% file: contents/02relatedwork.tex
\vspace{-1em}
\section{Related Work}
\vspace{-0.2em}
\subsection{Diffusion Models for Time Series}
Diffusion models have emerged as a powerful class of generative approaches, demonstrating remarkable success across various high-dimensional data domains. Denoising Diffusion Probabilistic Models (DDPM)~\cite{ho2020denoising} employ a Markov chain to add and remove Gaussian noise, producing high-fidelity samples gradually. Score-based diffusion models~\cite{song2020score} directly estimate the score function of data distributions. Meanwhile, conditional diffusion models incorporate additional guidance signals to enhance generation quality~\cite{dhariwal2021diffusion} further.
\begin{figure*}[t]
  \centering
  \includegraphics[width=0.98\linewidth]{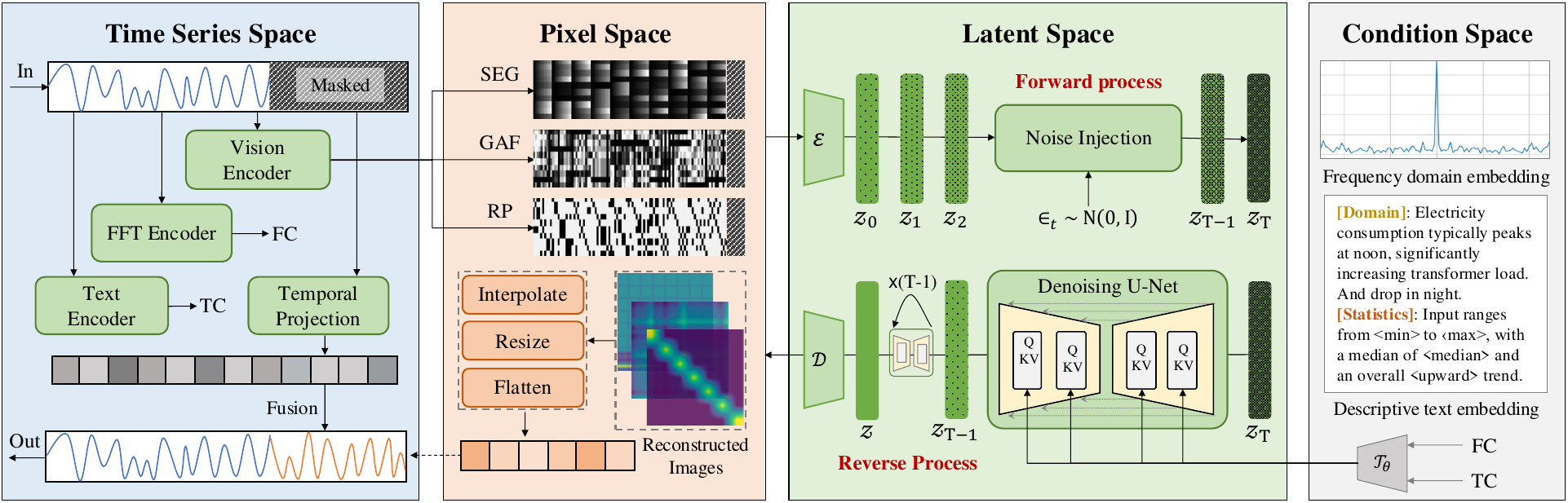}
  \caption{The framework of our proposed LDM4TS. Time series data is first transformed into complementary visual representations (SEG: Segmentation, GAF: Gramian Angular Field, RP: Recurrence Plot) that encode structural temporal patterns. A conditional latent diffusion model then reconstructs the masked images through iterative denoising guided by cross-modal conditioning (FC: frequency conditioning, TC: textual conditioning). Finally, the reconstructed images are mapped back to time series space with explicit temporal dependencies and implicit patterns.}
\vspace{-0.4cm}
\label{fig:framework}
\end{figure*}

Recent years have witnessed increasing applications of diffusion models in time series analysis~\cite{yang2024survey,lin2024diffusion}. For forecasting tasks, TimeGrad~\cite{rasul2021autoregressive} pioneers this direction by incorporating autoregressive components, while D3VAE~\cite{li2022generative} integrates variational auto-encoders with diffusion for enhanced capabilities. TSDiff~\cite{kollovieh2024predict} iteratively refines base forecasts through implicit probability densities. In parallel, conditional approaches have been explored for imputation tasks, where CSDI~\cite{tashiro2021csdi} and MIDM~\cite{wang2023observed} handle irregular time series through conditional score matching. Domain-specific designs have also emerged, such as DiffLoad~\cite{wang2023diffload} for load forecasting, WaveGrad~\cite{chen2020wavegrad} and DiffWave~\cite{kong2020diffwave} for audio synthesis, and EHRDiff~\cite{yuan2023ehrdiff} for healthcare applications. DiffSTG~\cite{wen2023diffstg} further explores spatio-temporal graph structures in diffusion models for time series.

\textit{However, these methods primarily focus on single-modality time series and lack mechanisms for modeling cross-modal temporal relationships.} Our work advances the development of latent diffusion models for TSF by incorporating multimodal information and exploiting cross-modal conditioning mechanisms, thereby substantially improving the accuracy and robustness under different forecasting scenarios.

\subsection{Vision-enhanced Time Series Forecasting}
Vision models like ViT~\cite{dosovitskiy2020image} and MAE~\cite{he2022masked} have demonstrated remarkable success in CV through their powerful feature extraction capabilities, showing strong generalization abilities when pre-trained on large-scale datasets like ImageNet~\cite{deng2009imagenet}. Inspired by the success of vision models, researchers have begun exploring their potential in time series forecasting. 

The concept of treating time series as images has evolved from traditional approaches using Gramian Angular Fields (GAF) and Markov Transition Fields (MTF)~\cite{wang2015imaging} to more sophisticated methods. TimesNet~\cite{wu2022timesnet} introduces a novel approach by transforming temporal data into 2D matrices, enabling the use of inception blocks for multi-scale temporal pattern extraction. Building upon this idea, SparseTSF~\cite{lin2024sparsetsf} incorporates sparsity constraints to better capture periodic and trend components in time series. ViTime~\cite{yang2024vitime} demonstrates the possibility of zero-shot forecasting by treating time series as visual signals while VisionTS~\cite{chen2024visionts} shows that pre-trained visual models can directly serve as time series forecasters without domain-specific adaptation. 

\textit{However, these approaches are predominantly deterministic and lack uncertainty quantification capabilities since they are not built within generative frameworks.} Our work addresses these limitations by integrating latent diffusion models with visual representations in a unified framework. This design enables our model to effectively capture temporal dependencies while maintaining the uncertainty modeling capabilities inherent in diffusion models.

\par In summary, while existing methods have made significant progress in either cross-modal temporal pattern extraction or probabilistic generative modeling, they have yet to effectively combine these complementary strengths. Our work bridges this gap by introducing a unified framework that leverages both visual representation learning and latent diffusion processes. Using multi-view vision transformations and multi-conditional generation, our approach captures rich temporal patterns while providing well-calibrated uncertainty estimates, advancing both visual representation learning and probabilistic modeling aspects of TSF.
\vspace{-0.5em}

%% file: contents/03method.tex
\section{Methodology}
\label{sec:method}
As illustrated in Figure~\ref{fig:framework}, \model employs a cross-modal architecture combining vision transformation and diffusion-based generation for time series processing. The framework first transforms raw time series data through complementary encoding methods, generating multi-view visual representations that capture diverse temporal patterns. These visual representations are then processed by a frozen latent diffusion model, which learns to reconstruct and predict temporal patterns under the guidance of multimodal conditioning. The model generates final predictions through a gated fusion mechanism that integrates the diffusion outputs with projected temporal features.

\subsection{Time Series to Multi-view Vision Transformation}
Time series data exhibits complex temporal patterns across multiple views, from local fluctuations to long-term trends, making direct modeling challenging. While existing methods predominantly rely on sequential architectures to capture temporal dependencies, they often fail to fully leverage the rich structural correlations embedded within time series data. To address these limitations, we propose a novel approach that transforms time series into visual representations, capturing multi-view temporal characteristics through a vision encoder (VE) and harnessing the sophisticated pattern recognition capabilities of latent diffusion models. Given an input sequence $X \in \mathbb{R}^{B \times L \times D}$, where $B$ denotes the batch size, $L$ represents the sequence length, and $D$ indicates the feature dimension, we construct a three-channel image representation through complementary encoding methods. The technical details of the complete transformation process are presented in Appendix~\ref{appx:transformation} and \ref{appx:visualization_picel_space}.

We transform time series data into three complementary visual representations, each designed to capture distinct temporal characteristics. Specifically, (i) the Segmentation representation (SEG)~\cite{chen2024visionts} that employs periodic restructuring to preserve local temporal structures, enabling the detection of recurring patterns across multiple time scales; (ii) the Gramian Angular Field (GAF)~\cite{zheng2014time,wang2015encoding} that transforms temporal correlations into spatial patterns through polar coordinate mapping, effectively capturing long-range dependencies crucial for forecasting; and (iii) the Recurrence Plot (RP)~\cite{eckmann1995recurrence,marwan2007recurrence} that constructs similarity matrices between time points to reveal both cyclical behaviors and temporal anomalies, providing a complementary view of the underlying structure. As demonstrated in Figure~\ref{fig:visual_cases}, these three visual encoding strategies effectively convert temporal dynamics into structured spatial patterns, enabling our model to capture local dependencies and global correlations through the diffusion process. The complete transformation process can be formalized as follows:
\begin{equation}
\small
\tilde{X} = \frac{X - \min(X)}{\max(X) - \min(X) + \epsilon}
\end{equation}
\begin{equation}
\small
I_{SEG} = \frac{1}{D}\sum_{d=1}^D \psi(f_{interp}(\mathcal{R}(\text{Pad}(\tilde{X}_{:,d,:}), \frac{L+p}{T}, T)))
\end{equation}
\begin{equation}
\small
I_{GAF} = f_{interp}(\frac{1}{D}\sum_{d=1}^D \cos(\theta_d \oplus \theta_d^T))
\end{equation}
\begin{equation}
\small
I_{RP} = f_{interp}(\exp(-\frac{\|X_i - X_j\|_2^2}{2}))
\end{equation}
\begin{equation} 
\small
I_m = \text{Concat}[I_{SEG}; I_{GAF}; I_{RP}]
\end{equation}
where $\mathcal{R}(X,m,n)$ transforms tensor $X$ into an $m \times n$ matrix for periodic pattern extraction, $\text{Pad}(\cdot)$ ensures sequence length divisibility by period $T$, $f_{interp}$ performs bilinear interpolation to target size $(H,W)$, $\psi(\cdot)$ normalizes each channel independently, and $\oplus$ denotes the outer sum operation. The resulting multi-channel image $I_m\in \mathbb{R}^{B \times 3 \times H \times W}$ integrates complementary views of temporal dynamics.

\subsection{Latent Diffusion for Time Series Reconstruction}
Unlike conventional diffusion models that operate in high-dimensional pixel space, we perform the denoising process in a compressed latent space, significantly reducing computational complexity while preserving temporal dynamics. Our framework extends Stable Diffusion~\cite{rombach2022high} with specialized adaptations for time series data through cross-modal conditional control and enhanced temporal modeling. The algorithm details are in Appendix~\ref{appx:ldm_algorithm}.
\paragraph{Multi-conditional Generation Framework}
To guide accurate temporal feature reconstruction, we devise a cross-modal conditioning mechanism that uses both frequency domain information and semantic descriptions. Given a visual representation $I \in \mathbb{R}^{B \times 3 \times H \times W}$, we first encode it into latent space and derive conditional signals as:
\begin{equation}
\small
    \quad c_{freq} = \text{FFTEncoder}(X),\quad c_{text} = \text{TextEncoder}(X)
\end{equation}
\begin{equation}
\small
    z = E(I) \cdot s,\quad c_m = \text{CrossAttn}(\text{MLP}([c_{text}; c_{freq}]), z)
\end{equation}
where $E(\cdot)$ represents the VAE encoder, $s$ is the latent space scaling factor (see Appendix~\ref{latent_space_scaling} for detailed derivation). $c_{freq}\in \mathbb{R}^{B \times (2DL+2)}$ captures periodic patterns through frequency analysis while $c_{text}\in \mathbb{R}^{B \times d_{model}}$ encodes statistical properties and domain knowledge through natural language descriptions. The detailed implementations of FFTEncoder and TextEncoder are provided in Appendix~\ref{appx:conditional_generation}.

\paragraph{Forward Diffusion Process}
The forward process implements a variance-preserving Markov chain that progressively injects Gaussian noise into the latent representations transformed from multi-view visual encodings of time series data. This controlled noise injection, operating in a compressed latent space rather than pixel space, enables efficient learning of temporal patterns across different scales while preserving the intrinsic information from vision transformations. For a given initial latent representation $z_0$, we define the forward diffusion process through a series of probabilistic transformations:
\begin{equation}
\small
    q(z_t|z_{t-1}, I_m) = \mathcal{N}(z_t; \sqrt{\alpha_t}z_{t-1}, (1-\alpha_t)I_m)
\end{equation}
\begin{equation}
\small
    q(z_t|z_0, I_m) = \mathcal{N}(z_t; \sqrt{\bar{\alpha}_t}E(I)/s, (1-\bar{\alpha}_t)I_m)
\end{equation}
\begin{equation}
\small
    \bar{\alpha}_t = \prod_{s=1}^t \alpha_s, \quad t \in \{1,...,T\}
\end{equation}
where $\{\alpha_t\}_{t=1}^T$ defines a scaled linear noise schedule, and $\bar{\alpha}_t$ controls the cumulative noise level across $t$ timesteps. The encoder $E(\cdot)$ maps the multi-view visual representations $I_m$ to a lower-dimensional latent space. 

\paragraph{De-noising Process}
The reverse process employs a parameterized U-Net architecture to progressively denoise the representations exploiting cross-modal conditioning mechanisms. By incorporating frequency and semantic embeddings, this process uniquely captures both complex temporal dynamics while maintaining coherent long-term dependencies. The complete denoising process is formulated as:
\begin{equation}
\small
    p_\theta(z_{t-1}|z_t,c_m) = \mathcal{N}(z_{t-1}; \mu_\theta(z_t,t,c_m), \Sigma_\theta(z_t,t))
\end{equation}
\begin{equation}
\small
    \mu_\theta(z_t,t,c_m) = \frac{1}{\sqrt{\alpha_t}}(z_t - \frac{1-\alpha_t}{\sqrt{1-\bar{\alpha}_t}}\epsilon_\theta(z_t,t,c_m))
\end{equation}
\begin{equation}
\small
    \mathcal{L} = \mathbb{E}_{z_0,\epsilon,t}[\|\epsilon - \epsilon_\theta(z_t,t,c_m)\|_2^2] + \lambda\mathcal{L}_{recon}
\end{equation}
where $\epsilon_\theta$ predicts the noise component given the noisy latent $z_t$, timestep $t$, and cross-modal condition $c_m$. The final reconstructed image $\hat{I} = D(z_0/s)$ is obtained by decoding the denoised latent representation. This design enables effective temporal pattern learning, comprehensive structural information preservation, and latent space optimization.
\begin{figure}[!t]
  \centering
  \includegraphics[width=0.95\linewidth]{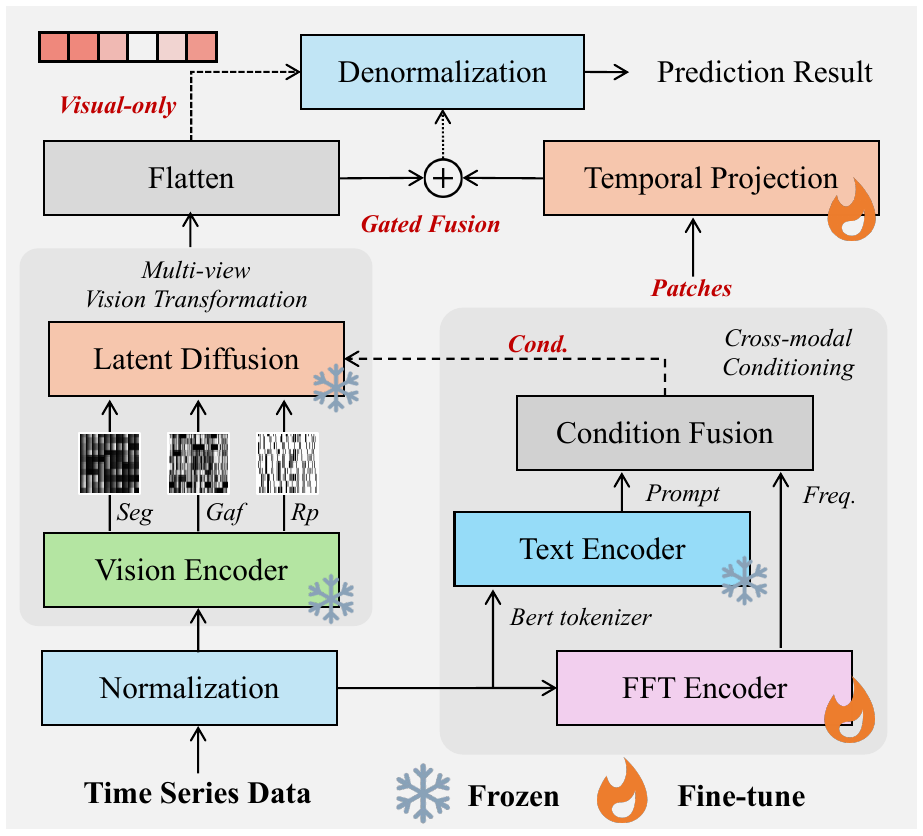}
  \caption{The forward process of LDM4TS.}
\label{fig:forward}
\end{figure}

\subsection{Temporal Projection and Fusion}
While the latent diffusion model captures global patterns effectively, local temporal dynamics and distribution shifts require explicit modeling. As shown in Fig.~\ref{fig:forward}, we propose a temporal encoder (TE) that complements the diffusion process through three key components: patch embedding, transformer encoding, and adaptive feature fusion.
Given input sequence $X \in \mathbb{R}^{B \times L \times D}$, we adopt the patch embedding strategy~\cite{dosovitskiy2020image,nie2023patchtst} to encode temporal information. Then, the embeddings are processed through $L$ transformer layers, where $X_{norm} = \text{LN}(X)$. The embeddings are processed as follows:
\begin{equation}
\small
    h_0 = \text{PatchEmbed}(X_{norm})\in \mathbb{R}^{B \times N_p \times d}
\end{equation}
\begin{equation}
\small
    h'_l = h_{l-1} + \text{MSA}(\text{LN}(h_{l-1}))
\end{equation}
\begin{equation}
\small
    h_l = h'_l + \text{MLP}(\text{LN}(h'_l))
\end{equation}
\begin{equation}
\small
    Z_{TE} = \text{LinearProj}(h_L) \in \mathbb{R}^{B \times L_{pred} \times D}
\end{equation}
where $N_p$ denotes patch count, $h$ is hidden states and $d$ is the hidden dimension. MSA and LN represent multi-head self-attention and layer normalization respectively. Finally, we implement an adaptive fusion mechanism that dynamically combines temporal features $Z_{TE}$ with visual features $Z_{VE}$:
\begin{equation}
\small
    g = \text{Sigmoid}(\text{MLP}([Z_{TE}; Z_{VE}])) 
\end{equation}
\begin{equation}
\small
    \hat{Y} = g \odot Z_{TE} + (1-g) \odot Z_{VE}
\end{equation}
where learned gates $g$ dynamically balance the contribution of global patterns from diffusion and local dynamics from temporal encoding, enabling robust adaptation to different temporal characteristics.

%% file: contents/04experiments.tex
\input{table/long_term_forecasting}
\input{table/few_shot}

\section{Experiments}
\label{sec:experiments}
\subsection{Settings}
\paragraph{Dataset and Metrics} 
In this section, we evaluate the proposed \model on seven widely-used time series datasets, covering diverse domains including energy consumption (ETTh1, ETTh2, ETTm1, ETTm2), weather forecasting, electricity load prediction (ECL, 321 variables), and traffic flow estimation (Traffic, 862 variables)~\cite{zhou2021informer, lai2018modeling}, which have been extensively adopted for benchmarking long-term forecasting models~\cite{wu2022timesnet}. These datasets are chosen for their varying characteristics in terms of sampling frequency, dimensionality, and temporal patterns. Forecasting performance is measured using Mean Absolute Error (MAE) and Mean Squared Error (MSE), following standard practice in the field. Details are provided in Appendix~\ref{appx:dataset_details} and~\ref{appx:metric}.
\paragraph{Compared Methods} 
We compare with 1) transformer-based methods: 
FEDformer~\cite{zhou2022fedformer}, Autoformer~\cite{wu2021autoformer}, Informer~\cite{zhou2021informer}, 
ETSformer~\cite{woo2022etsformer}, and Reformer~\cite{kitaev2020reformer}. 
2) diffusion-based methods: 
CSDI~\cite{tashiro2021csdi} and ScoreGrad~\cite{yan2021scoregrad}. 
3) a set of recent competitive models, including 
DLinear~\cite{zeng2023transformers}, TimesNet~\cite{wu2022timesnet}, 
and LightTS~\cite{zhang2022less}. 
These baselines represent state-of-the-art approaches in time series forecasting, encompassing different methodological paradigms from probabilistic generative modeling to attention-based architectures and linear models. More details are in Appendix~\ref{appx:baselines}.

\paragraph{Implementation Details} 
The models are trained using the Adam optimizer with a learning rate of $10^{-3}$, batch size of $32$, and a maximum of $10$ epochs, applying an early stopping strategy. The number of diffusion steps is set to $T=300$, with a linear variance schedule from $\beta_1=0.00085$ to $\beta_K=0.012$. The validation set determines the history window length (selected from $\{96, 168, 336, 720\}$). All experiments are conducted on an Nvidia RTX A6000 GPU with 48GB memory. All training and model parameter settings with default values are detailed in Appendix~\ref{appx:optimization_settings}.
\subsection{Long-term Forecasting}
We evaluate the long-term forecasting capabilities of \model across multiple prediction horizons. As shown in Table~\ref{tab:long_term_results}, \model consistently outperforms state-of-the-art baselines, achieving optimal results in both MSE and MAE. On the ETT dataset family, our approach demonstrates significant improvements, achieving the best MSE of 0.352 on ETTm1 compared to the second-best performer DLinear (0.404), and reducing MSE by 11.8\% on ETTh2 (0.387) compared to FEDformer (0.439). The advantages extend to high-dimensional scenarios, achieving superior results on both Electricity (321 variables, MSE: 0.199 vs TimesNet 0.208) and Traffic datasets (862 variables, MSE: 0.550 vs DLinear 0.624), which is mainly benefited by our latent diffusion framework that compresses high-dimensional temporal patterns into a lower-dimensional latent space while preserving essential structural information through vision transformations. Notably, \model substantially outperforms existing diffusion-based methods CSDI and ScoreGrad across all datasets, with MSE improvements of up to 84.2\% and 89.4\% on the Traffic dataset respectively. Overall, \model achieves the best performance in 5 out of 7 datasets on each metric, validating that our vision-enhanced modeling strategy effectively captures complex temporal dynamics across diverse forecasting scenarios. 

\subsection{Few-shot Forecasting}
To evaluate model robustness under data scarcity, we conduct experiments using only 10\% and 5\% of the training data. As shown in Table~\ref{tab:few_shot}, \model achieves optimal performance on 6 out of 7 datasets in both MSE and MAE metrics. On the ETT benchmark series, \model consistently outperforms state-of-the-art methods, with notable MSE reductions: 26.2\% on ETTh1 (0.471 vs 0.638), 3.2\% on ETTh2 (0.452 vs 0.466), and 9.7\% on ETTm1 (0.371 vs 0.411). The advantages extend to high-dimensional scenarios, with \model outperforming DLinear by 4.4\% on Electricity (0.172 vs 0.180) and FEDformer by 6.3\% on Traffic (0.621 vs 0.663). Notably, \model shows significant improvements over diffusion-based methods CSDI and ScoreGrad, reducing average MSE by 25.6\% and 31.2\% respectively. Even with further reduced 5\% training data, \model maintains strong performance by achieving the best results on 5 MSE and 6 MAE metrics across datasets. The consistently robust performance under extreme data scarcity demonstrates how our vision-enhanced approach captures intrinsic patterns to address fundamental challenges in real-world forecasting applications.

\input{table/few_shot_5p}
\input{table/zero_shot}

\subsection{Zero-Shot Forecasting}
To evaluate cross-domain generalization, we conduct zero-shot transfer experiments across different datasets without any fine-tuning. As shown in Table~\ref{tab:zero_shot_results}, \model achieves the best performance in 4 MSE and 5 MAE metrics out of 8 scenarios, demonstrating strong cross-domain transferability. For challenging transfer tasks like $ETTh1\rightarrow ETTh2$ and $ETTh1\rightarrow ETTm2$, \model achieves MSE of 0.458 and 0.369 respectively, outperforming both DLinear (0.493, 0.415) and FEDformer (0.495, 0.373). The model also achieves the best of on $ETTm1\rightarrow ETTh1$ (0.452, 0.434) and $ETTm2\rightarrow ETTm1$ (0.588, 0.487). The advantages are particularly pronounced when compared to diffusion models, with \model achieving substantial improvements over both CSDI and ScoreGrad across all transfer scenarios, reducing MSE by up to 49.0\% in challenging tasks. Notably, while most baseline methods show significant performance degradation in cross-dataset transfers, \model maintains consistent performance across different transfer pairs, suggesting robust generalization capabilities. 


\subsection{Model Analysis}
\begin{figure}[!ht]
  \centering
\includegraphics[width=\linewidth]{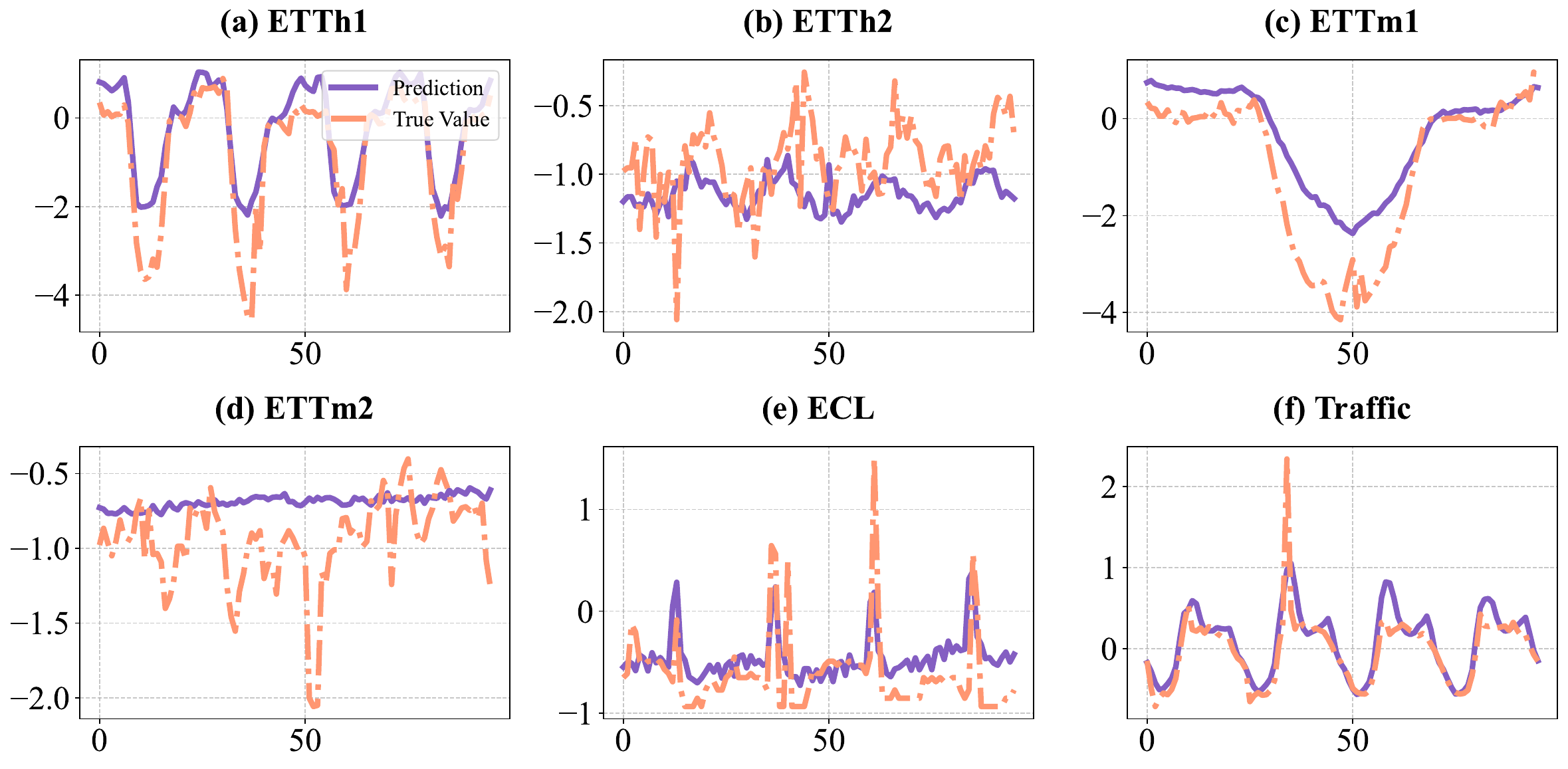}
  \caption{Visualization results of long-term forecasting by LDM4TS model on all datasets under the input-96-predict-96 setting. Detailed comparisons with baselines on the ETTh1 dataset are in the Appendix~\ref{appx:showcases}.}
\vspace{-2mm}
\label{fig:vis_forecast}
\end{figure}

\paragraph{Overall Performance Analysis}
\model demonstrates superior performance across various forecasting scenarios, excelling in long-term few-shot, and zero-shot predictions, while maintaining computational efficiency with only 5.4M learnable parameters and fast inference speed (see Append~\ref{appx:efficiency} for detailed analysis). Through comprehensive experiments, we observe that our approach effectively captures both global trends and local patterns in time series data. As shown in Figure~\ref{fig:vis_forecast}, \model achieves good performance in forecasting structured patterns, such as the clear periods in Traffic datasets (MSE: 0.621) and regular consumption patterns in ECL data (MSE: 0.199). The performance shows slight degradation on datasets with irregular patterns or abrupt changes, suggesting potential areas for future improvement in handling non-stationary patterns.

\begin{figure}[!ht]
  \centering
  \includegraphics[width=0.95\linewidth]{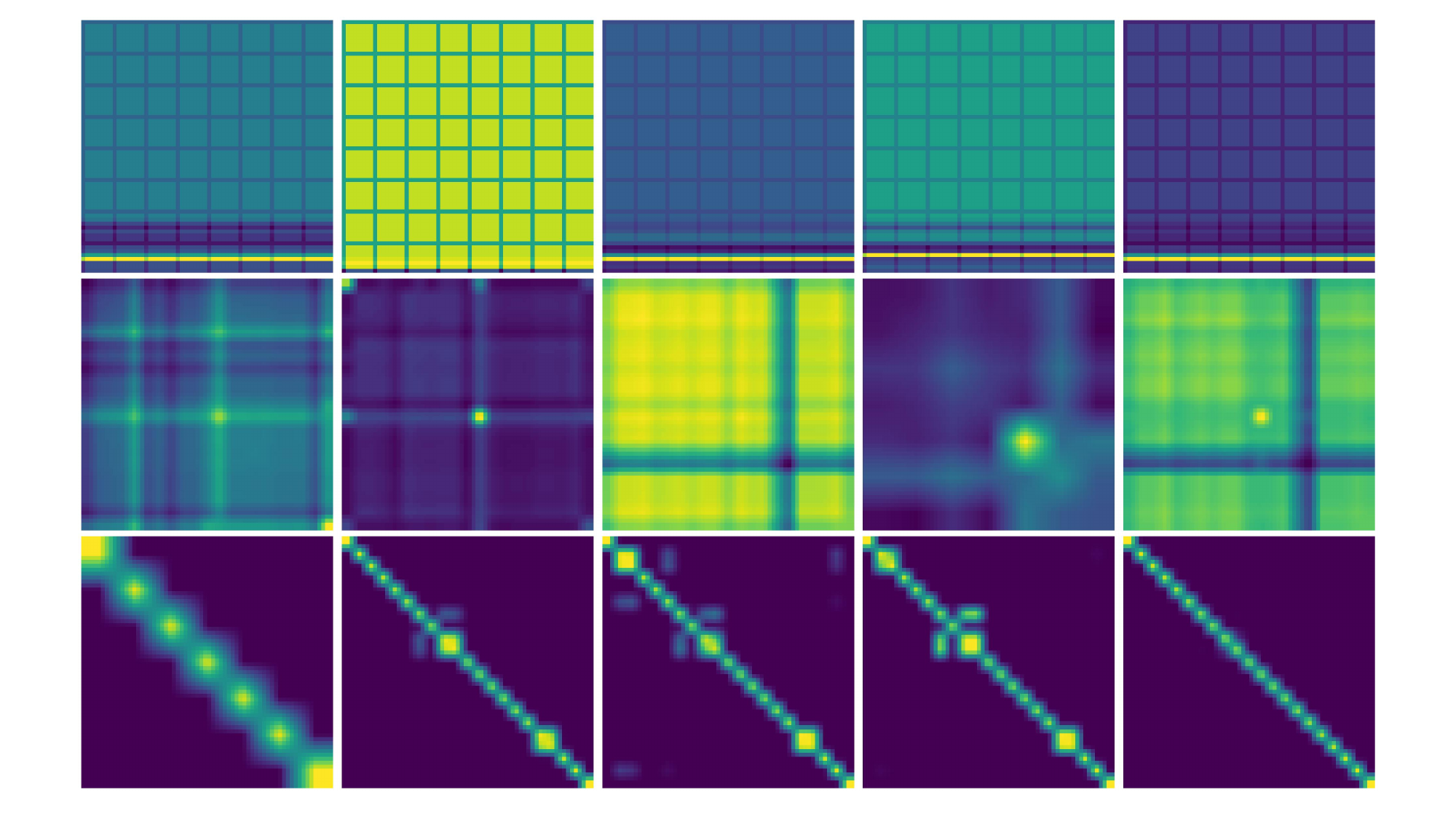}
  \caption{Visualization of multi-view visual representation after transformation. Each row shows one approach, top row: Segmentation (SEG); middle row: Gramian Angular Field (GAF); and bottom row: Recurrence Plot (RP). More detailed results are provided in Appendix~\ref{appx:visualization_picel_space}.}
\vspace{-1mm}
\label{fig:visual_cases}
\end{figure}

\input{table/ablation_components}
\paragraph{Visual Encoding Effectiveness}
Figure~\ref{fig:visual_cases} illustrates our multi-view vision transformation strategy, which forms the foundation of \model's strong performance. Each encoding method captures distinct temporal characteristics: SEG preserves local temporal structures through periodic restructuring, enabling detection of recurring patterns at multiple time scales; GAF transforms temporal correlations into spatial patterns through polar coordinate mapping, effectively capturing long-range dependencies; and RP generates similarity matrices that highlight both cyclical behaviors and temporal anomalies. The complementary nature of these encodings is particularly evident in the ETT datasets, where the combination achieves a 24.2\% reduction in MSE compared to using any single encoding method.

\begin{figure}[!ht]
  \centering
  \includegraphics[width=\linewidth]{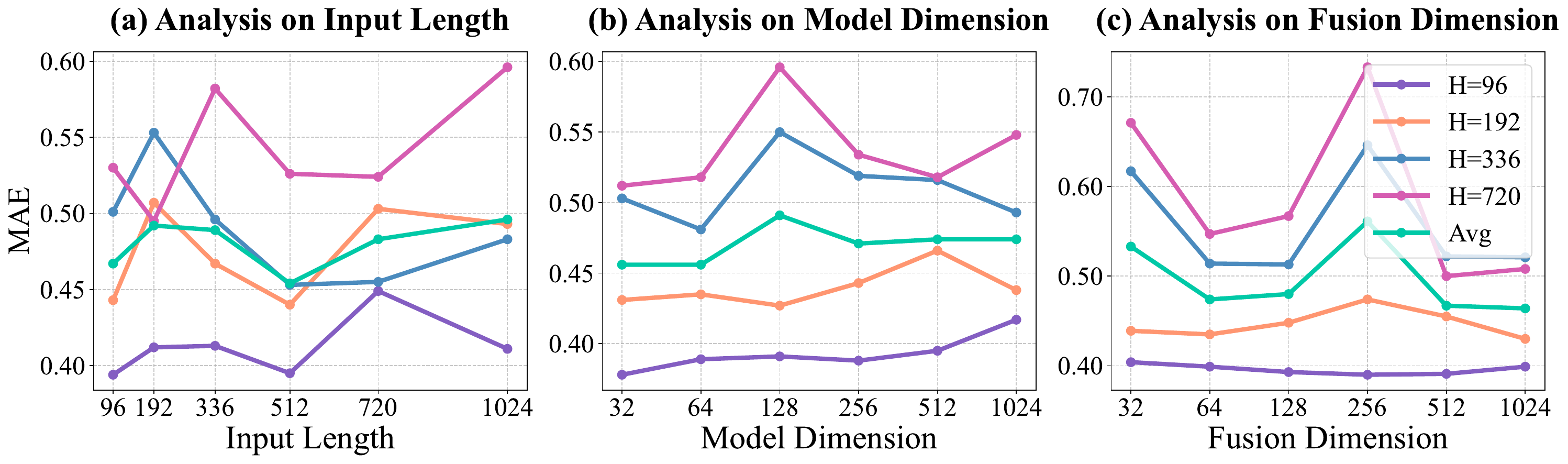}
  \caption{Hyperparameter sensitivity analysis. The impact of sequence length (left), model dimension (middle), and fusion dimension (right) on prediction performance on the ETTh1 dataset.}
\label{fig:vis_parameter}
\vspace{-1mm}
\end{figure}
\paragraph{Ablation Studies}
Table~\ref{tab:ablation_components} presents ablation studies on key components of \model. Both vision encoder and temporal encoder prove to be crucial, with their removal leading to significant performance degradation (10.57\% and 10.96\% MSE increase respectively), validating that our visual representations successfully capture and align with temporal characteristics. The latent diffusion module also plays a vital role (5.57\% MSE increase when removed), demonstrating effective bridging between image reconstruction and time series prediction. Frequency conditioning shows minimal impact (0.59\% MSE increase) due to information redundancy. Unexpectedly, removing the text conditioning resulted in a performance drop, suggesting the textual information may lack an underlying temporal statistical pattern.
We further performed a parameter sensitivity analysis to investigate the effect of key hyperparameters on the model performance, as shown in Figure~\ref{fig:vis_parameter}(a) shows the best performance at around 512 timesteps as input sequence length, while the performance of longer sequences decreases due to increased noise. The model hidden dimension shows an optimum point between 32 and 64, balancing model capacity and risk of overfitting. For the hidden dimension of the fusion module, values between 64 and 128 produce better results, suggesting that compact representations are more effective for integrating cross-modal information.

%% file: table/long_term_forecasting.tex
\begin{table*}[!ht]
\setlength{\tabcolsep}{2pt}
\captionsetup{font=small}
  \centering
  \caption{Long-term forecasting results. 
All results are averaged from four forecasting horizons:  $H \in \{96, 192, 336, 720\}$. A lower value indicates better performance. \boldres{Red}: best, \secondres{Blue}: second best. Our full results are in Appendix~\ref{appx:long_term_details}.}
\scalebox{0.95}{
    \begin{tabular}{c|cc|cc|cc|cc|cc|cc|cc|cc|cc|cc}
    \toprule
    \multirow{2}{*}{Methods} & \multicolumn{2}{c|}{LDM4TS} & \multicolumn{2}{c|}{CSDI} & \multicolumn{2}{c|}{ScoreGrad} & \multicolumn{2}{c|}{Autoformer} & \multicolumn{2}{c|}{FEDformer} & \multicolumn{2}{c|}{DLinear} & \multicolumn{2}{c|}{Informer} & \multicolumn{2}{c|}{TimesNet} & \multicolumn{2}{c|}{LightTS} & \multicolumn{2}{c}{Reformer} \\
    & \multicolumn{2}{c|}{\textbf{(Ours)}} & \multicolumn{2}{c|}{\citeyearpar{tashiro2021csdi}} & \multicolumn{2}{c|}{\citeyearpar{song2020score}} & \multicolumn{2}{c|}{\citeyearpar{wu2021autoformer}} & \multicolumn{2}{c|}{\citeyearpar{zhou2022fedformer}} & \multicolumn{2}{c|}{\citeyearpar{zeng2023transformers}} & \multicolumn{2}{c|}{\citeyearpar{zhou2021informer}} & \multicolumn{2}{c|}{\citeyearpar{wu2022timesnet}} & \multicolumn{2}{c}{\citeyearpar{campos2023lightts}} & \multicolumn{2}{c}{\citeyearpar{kitaev2020reformer}}\\
    \midrule
    Metric & MSE & MAE & MSE & MAE & MSE & MAE & MSE & MAE & MSE & MAE & MSE & MAE & MSE & MAE & MSE & MAE & MSE & MAE & MSE & MAE \\
    \midrule
    \textit{ETTh1} & \textcolor{blue}{\underline{0.443}} & \textcolor{red}{\textbf{0.454}} & 0.987 & 0.742 & 1.031 & 0.709 & 0.504 & 0.492 & \textcolor{red}{
    \textbf{0.439}} & 0.458 & 0.460 & 0.457 & 1.040 & 0.799 & 0.460 & \textcolor{blue}{\underline{0.455}} & 0.590 & 0.544 &1.006 & 0.745\\
    \textit{ETTh2} & \textcolor{red}{\textbf{0.387}} & \textcolor{blue}{\underline{0.427}} & 0.539 & 0.520 & 0.512 & 0.505 & 0.467 & 0.468 & 0.439 & 0.451 & 0.564 & 0.519 & 4.551 & 1.742 & \textcolor{blue}{\underline{0.407}} & \textcolor{red}{
    \textbf{0.421}} & 1.260 & 0.678 & 2.531 & 1.244\\
    \textit{ETTm1} & \textcolor{red}{\textbf{0.352}} & \textcolor{red}{\textbf{0.387}} & 0.806 & 0.648 & 1.016 & 0.678 & 0.576 & 0.526 & 0.471 & 0.470 & \textcolor{blue}{\underline{0.404}} & \textcolor{blue}{\underline{0.408}} & 0.867 & 0.690 & 0.477 & 0.443 & 0.427 & 0.437 & 1.013 & 0.737\\
    \textit{ETTm2} & 0.333 & 0.380 & 0.381 & 0.425 & 0.446 & 0.447 & 0.307 & 0.351 & 0.318 & 0.366 & \textcolor{blue}{\underline{0.304}} & \textcolor{blue}{\underline{0.349}} & 1.593 & 0.908 & \textcolor{red}{\textbf{0.299}} & \textcolor{red}{\textbf{0.333}} & 0.830 & 0.614 & 1.874 & 1.009 \\
    \textit{Weather} & \textcolor{red}{\textbf{0.245}} & \textcolor{red}{\textbf{0.283}} & 0.296 & 0.335 & 0.347 & 0.351 & 0.329 & 0.375 & 0.333 & 0.375 & \textcolor{red}{\textbf{0.245}} & 0.306 & 0.634 & 0.549 & 0.265 & \textcolor{blue}{\underline{0.288}} & \textcolor{blue}{\underline{0.259}} & 0.315 & 1.229 & 0.858\\
    \textit{ECL} & \textcolor{red}{\textbf{0.199}} & \textcolor{red}{\textbf{0.299}} & 0.993 & 0.804 & 1.258 & 0.884 & 0.253 & 0.352 & 0.612 & 0.377 & 0.225 & 0.319 & 0.378 & 0.438 & \textcolor{blue}{\underline{0.208}} & \textcolor{blue}{\underline{0.303}} & 0.243 & 0.343 & 0.326 & 0.404\\
    \textit{Traffic} & \textcolor{red}{\textbf{0.550}} & \textcolor{red}{\textbf{0.321}} & 1.736 & 0.869 & 2.099 & 1.020 & 0.641 & 0.492 & 0.612 & 0.377 & 0.624 & 0.384 & 0.832 & 0.465 & 0.636 & \textcolor{blue}{\underline{0.336}} & \textcolor{blue}{\underline{0.609}} & 0.394 & 0.717 & 0.399\\
\midrule
    1st Count & 5 & 5 & 0 & 0 & 0 & 0 & 0 & 0 & 1 & 0 & 1 & 0 & 0 & 0 & 1 & 2 & 0 & 0 & 0 & 0\\
    \bottomrule
    \end{tabular}%
}
  \label{tab:long_term_results}%
\end{table*}%

%% file: table/few_shot.tex
\begin{table*}[!ht]
\setlength{\tabcolsep}{2pt}
\captionsetup{font=small}
  \centering
  \caption{Few-shot learning on 10\% training data. We use the same protocol in Table~\ref{tab:long_term_results}. All results are averaged from four different forecasting horizons: $H \in \{96, 192, 336, 720\}$. \boldres{Red}: best, \secondres{Blue}: second best. Our full results are in Appendix~\ref{appx:few_shot_details}.}
\scalebox{0.95}{
    \begin{tabular}{c|cc|cc|cc|cc|cc|cc|cc|cc|cc|cc}
    \toprule
    \multirow{2}{*}{Methods} & \multicolumn{2}{c|}{LDM4TS} & \multicolumn{2}{c|}{CSDI} & \multicolumn{2}{c|}{ScoreGrad} & \multicolumn{2}{c|}{Autoformer} & \multicolumn{2}{c|}{FEDformer} & \multicolumn{2}{c|}{DLinear} & \multicolumn{2}{c|}{Informer} & \multicolumn{2}{c|}{TimesNet} & \multicolumn{2}{c|}{LightTS} & \multicolumn{2}{c}{Reformer} \\
    & \multicolumn{2}{c|}{\textbf{(Ours)}} & \multicolumn{2}{c|}{\citeyearpar{tashiro2021csdi}} & \multicolumn{2}{c|}{\citeyearpar{song2020score}} & \multicolumn{2}{c|}{\citeyearpar{wu2021autoformer}} & \multicolumn{2}{c|}{\citeyearpar{zhou2022fedformer}} & \multicolumn{2}{c|}{\citeyearpar{zeng2023transformers}} & \multicolumn{2}{c|}{\citeyearpar{zhou2021informer}} & \multicolumn{2}{c|}{\citeyearpar{wu2022timesnet}} & \multicolumn{2}{c|}{\citeyearpar{campos2023lightts}} & \multicolumn{2}{c}{\citeyearpar{kitaev2020reformer}}\\
    \midrule
    Metric & MSE & MAE & MSE & MAE & MSE & MAE & MSE & MAE & MSE & MAE & MSE & MAE & MSE & MAE & MSE & MAE & MSE & MAE & MSE & MAE \\
    \midrule
    \textit{ETTh1} & \textcolor{red}{\textbf{0.471}} & \textcolor{red}{\textbf{0.468}} & 0.849 & 0.665 & 1.031 & 0.709 & 0.701 & 0.596 & \textcolor{blue}{\underline{0.638}} & \textcolor{blue}{\underline{0.561}} & 0.691 & 0.599 & 1.199 & 0.808 & 0.869 & 0.628 & 1.375 & 0.877 & 1.249 & 0.833\\
    \textit{ETTh2} & \textcolor{red}{\textbf{0.452}} & \textcolor{red}{\textbf{0.460}} & 0.527 & 0.523 & 0.512 & 0.505 & 0.488 & 0.499 & \textcolor{blue}{\underline{0.466}} & 0.475 & 0.608 & 0.538 & 3.871 & 1.512 & 0.479 & \textcolor{blue}{\underline{0.465}} & 2.655 & 1.159 & 3.485 & 1.486\\
    \textit{ETTm1} & \textcolor{red}{\textbf{0.371}} & \textcolor{red}{\textbf{0.393}} & 0.784 & 0.606 & 1.015 & 0.678 & 0.802 & 0.628 & 0.721 & 0.605 & \textcolor{blue}{\underline{0.411}} & \textcolor{blue}{\underline{0.429}} & 1.192 & 0.820 & 0.479 & 0.465 & 0.970 & 0.704 & 1.426 & 0.856 \\
    \textit{ETTm2} & 0.336 & 0.373 & 0.334 & 0.385 & 0.446 & 0.447 & 1.341 & 0.930 & 0.463 & 0.488 & \textcolor{red}{\textbf{0.316}} & \textcolor{blue}{\underline{0.368}} & 3.369 & 1.439 & \textcolor{blue}{\underline{0.319}} & \textcolor{red}{\textbf{0.353}} & 0.987 & 0.755 & 3.978 & 1.587 \\
    
    \textit{Weather} & \textcolor{red}{\textbf{0.229}} & \textcolor{red}{\textbf{0.276}} & 0.295 & 0.333 & 0.347 & 0.351 & 0.300 & 0.342 & 0.284 & \textcolor{blue}{\underline{0.283}} & \textcolor{blue}{\underline{0.241}} & \textcolor{blue}{\underline{0.283}} & 0.597 & 0.494 & 0.279 & 0.301 & 0.289 & 0.322 & 0.526 & 0.469\\
    
    \textit{ECL} & \textcolor{red}{\textbf{0.172}} & \textcolor{red}{\textbf{0.275}} & 0.909 & 0.785 & 1.258 & 0.884 & 0.431 & 0.478 & 0.346 & 0.428 & \textcolor{blue}{\underline{0.180}} & \textcolor{blue}{\underline{0.280}} & 1.194 & 0.891 & 0.323 & 0.392 & 0.441 & 0.488 & 0.980 & 0.769 \\
    
    \textit{Traffic} & \textcolor{red}{\textbf{0.621}} & \textcolor{red}{\textbf{0.357}} & 1.744 & 0.871 & 2.100 & 1.020 & 0.749 & 0.446 & \textcolor{blue}{\underline{0.663}} & \textcolor{blue}{\underline{0.425}} & 0.945 & 0.570 & 1.534 & 0.811 & 0.951 & 0.535 & 1.247 & 0.684 & 1.551 & 0.821\\
\midrule
    1st Count & 6 & 6 & 0 & 0 & 0 & 0 & 0 & 0 & 0 & 0 & 1 & 0 & 0 & 0 & 0 & 1 & 0 & 0 & 0 & 0\\
    \bottomrule
    \end{tabular}%
}
  \label{tab:few_shot}%
\end{table*}%

%% file: table/few_shot_5p.tex
\begin{table*}[!ht]
\setlength{\tabcolsep}{2pt}
\captionsetup{font=small}
  \centering
  \caption{Few-shot learning on 5\% training data. \boldres{Red}: best, \secondres{Blue}: second best. The full results are in Appendix~\ref{appx:few_shot_details}}
\scalebox{0.95}{
    \begin{tabular}{c|cc|cc|cc|cc|cc|cc|cc|cc|cc|cc}
    \toprule
    \multirow{2}{*}{Methods} & \multicolumn{2}{c|}{LDM4TS} & \multicolumn{2}{c|}{CSDI} & \multicolumn{2}{c|}{ScoreGrad} & \multicolumn{2}{c|}{Autoformer} & \multicolumn{2}{c|}{FEDformer} & \multicolumn{2}{c|}{DLinear} & \multicolumn{2}{c|}{Informer} & \multicolumn{2}{c|}{TimesNet} & \multicolumn{2}{c|}{LightTS} & \multicolumn{2}{c}{Reformer} \\
    & \multicolumn{2}{c|}{\textbf{(Ours)}} & \multicolumn{2}{c|}{\citeyearpar{tashiro2021csdi}} & \multicolumn{2}{c|}{\citeyearpar{song2020score}} & \multicolumn{2}{c|}{\citeyearpar{wu2021autoformer}} & \multicolumn{2}{c|}{\citeyearpar{zhou2022fedformer}} & \multicolumn{2}{c|}{\citeyearpar{zeng2023transformers}} & \multicolumn{2}{c|}{\citeyearpar{zhou2021informer}} & \multicolumn{2}{c|}{\citeyearpar{wu2022timesnet}} & \multicolumn{2}{c|}{\citeyearpar{campos2023lightts}} & \multicolumn{2}{c}{\citeyearpar{kitaev2020reformer}}\\
    \midrule
    Metric & MSE & MAE & MSE & MAE & MSE & MAE & MSE & MAE & MSE & MAE & MSE & MAE & MSE & MAE & MSE & MAE & MSE & MAE & MSE & MAE \\
    \midrule
\textit{ETTh1} & \boldres{0.458} & \boldres{0.456} & 0.850 & 0.662 & 1.026 & 0.687 & 0.722 & 0.599 & \secondres{0.659} & \secondres{0.562} & 0.750 & 0.611 & 1.225 & 0.817 & 0.926 & 0.648 & 1.451 & 0.903 & 1.242 & 0.835 \\
\textit{ETTh2} & 0.496 & 0.470 & 0.522 & 0.522 & 0.502 & 0.477 & 0.470 & 0.489 & \boldres{0.441} & \secondres{0.457} & 0.828 & 0.616 & 3.923 & 1.654 & \secondres{0.464} & \boldres{0.454} & 3.206 & 1.268 & 3.527 & 1.473 \\
\textit{ETTm1} & \secondres{0.407} & \boldres{0.412} & 0.784 & 0.603 & 1.009 & 0.659 & 0.796 & 0.621 & 0.731 & 0.593 & \boldres{0.401} & \secondres{0.417} & 1.163 & 0.791 & 0.717 & 0.561 & 1.123 & 0.766 & 1.264 & 0.827 \\
\textit{ETTm2} & \boldres{0.311} & \boldres{0.353} & \secondres{0.335} & 0.387 & 0.379 & 0.400 & 0.388 & 0.433 & 0.381 & 0.404 & 0.399 & 0.426 & 3.658 & 1.489 & 0.345 & \secondres{0.373} & 1.416 & 0.871 & 3.582 & 1.487 \\
\textit{Weather} & \boldres{0.258} & \boldres{0.294} & 0.295 & 0.333 & 0.347 & 0.351 & 0.311 & 0.354 & 0.310 & 0.353 & \secondres{0.264} & \secondres{0.309} & 0.584 & 0.528 & 0.298 & 0.318 & 0.306 & 0.345 & 0.447 & 0.453 \\
\textit{ECL} & \boldres{0.208} & \boldres{0.301} & 0.916 & 0.786 & 1.245 & 0.877 & 0.346 & 0.405 & 0.267 & 0.353 & \secondres{0.246} & \secondres{0.342} & 1.281 & 0.930 & 0.402 & 0.453 & 0.878 & 0.726 & 1.289 & 0.904 \\
\textit{Traffic} & \boldres{0.632} & \boldres{0.366} & 1.729 & 0.870 & 2.092 & 1.019 & 0.833 & 0.502 & \secondres{0.677} & \secondres{0.424} & 0.696 & 0.433 & 1.591 & 0.832 & 0.867 & 0.493 & 1.557 & 0.796 & 1.619 & 0.851 \\
\midrule
1st Count & 5 & 6 & 0 & 0 & 0 & 0 & 0 & 0 & 1 & 0 & 1 & 0 & 0 & 0 & 0 & 1 & 0 & 0 & 0 & 0\\
    \bottomrule
    \end{tabular}%
}
  \label{tab:few_shot_5p}%
\end{table*}%

%% file: table/zero_shot.tex
\begin{table*}[!htbp]
\setlength{\tabcolsep}{2pt}
\captionsetup{font=small}
  \centering
  \caption{Zero-shot learning results. All results are averaged from four forecasting horizons:  $H \in \{96, 192, 336, 720\}$. \boldres{Red}: best, \secondres{Blue}: second best. Appendix~\ref{appx:zero_shot_details} shows our detailed results.}
\scalebox{0.95}{
    \begin{tabular}{c|cc|cc|cc|cc|cc|cc|cc|cc|cc|cc}
    \toprule
    \multirow{2}{*}{Methods} & \multicolumn{2}{c|}{LDM4TS} & \multicolumn{2}{c|}{CSDI} & \multicolumn{2}{c|}{ScoreGrad} & \multicolumn{2}{c|}{Autoformer} & \multicolumn{2}{c|}{FEDformer} & \multicolumn{2}{c|}{DLinear} & \multicolumn{2}{c|}{Informer} & \multicolumn{2}{c|}{ETSformer} & \multicolumn{2}{c|}{LightTS} & \multicolumn{2}{c}{Reformer}\\
    & \multicolumn{2}{c|}{\textbf{(Ours)}} & \multicolumn{2}{c|}{\citeyearpar{tashiro2021csdi}} & \multicolumn{2}{c|}{\citeyearpar{song2020score}} & \multicolumn{2}{c|}{\citeyearpar{wu2021autoformer}} & \multicolumn{2}{c|}{\citeyearpar{zhou2022fedformer}} & \multicolumn{2}{c|}{\citeyearpar{zeng2023transformers}} & \multicolumn{2}{c|}{\citeyearpar{zhou2021informer}} & \multicolumn{2}{c|}{\citeyearpar{woo2022etsformer}} & \multicolumn{2}{c|}{\citeyearpar{campos2023lightts}} & \multicolumn{2}{c}{\citeyearpar{kitaev2020reformer}} \\
    \midrule
    Metric & MSE & MAE & MSE & MAE & MSE & MAE & MSE & MAE & MSE & MAE & MSE & MAE & MSE & MAE & MSE & MAE & MSE & MAE & MSE & MAE\\
    \midrule
    \textit{ETTh1$\rightarrow$ETTh2} & \textcolor{red}{\textbf{0.458}} & \textcolor{red}{\textbf{0.452}} & 0.500 & 0.527 & 0.512 & 0.505 & 0.582 & 0.548 & 0.495 & 0.501 & \textcolor{blue}{\underline{0.493}} & \textcolor{blue}{\underline{0.488}} & 2.292 & 1.169 & 0.589	& 0.589 & 1.075 & 0.699 & 2.119 & 1.125\\
    
    \textit{ETTh1$\rightarrow$ETTm2} & \textcolor{red}{\textbf{0.369}} & \textcolor{red}{\textbf{0.400}} & 0.410 & 0.444 & 0.446 & 0.447 & 0.457 & 0.483 & \textcolor{blue}{\underline{0.373}} & \textcolor{blue}{\underline{0.424}} & 0.415 & 0.452 & 2.167 & 1.124 & 0.569 & 0.568 & 1.058 & 0.700 & 2.228 & 1.165\\
    
    \textit{ETTh2$\rightarrow$ETTh1} & 0.723 & 0.577 & 1.416 & 0.898 & 1.032 & 0.709 & 0.757 & 0.608 & 1.423 & 0.803 & \textcolor{blue}{\underline{0.703}} & \textcolor{blue}{\underline{0.574}} & 1.713 & 0.978 & 0.864	& 0.675 & \textcolor{red}{\textbf{0.567}} & \textcolor{red}{\textbf{0.518}} & 0.909 & 0.705 \\
    
    \textit{ETTh2$\rightarrow$ETTm2} & \textcolor{blue}{\underline{0.432}} & \textcolor{blue}{\underline{0.444}} & 0.397 & 0.437 & 0.446 & 0.447 & \textcolor{red}{\textbf{0.386}} & \textcolor{red}{\textbf{0.411}} & 0.433 & 0.460 & 0.328 & 0.386 & 4.606 & 1.763 & 1.320 & 0.917 & 0.703	& 0.585 & 2.726 & 1.257\\
    
    \textit{ETTm1$\rightarrow$ETTh2} & \textcolor{red}{\textbf{0.452}} & \textcolor{red}{\textbf{0.434}} & 0.504 & 0.515 & 0.505 & 0.505 & \textcolor{blue}{\underline{0.470}} & \textcolor{blue}{\underline{0.479}} & 0.587 & 0.565 & 0.464 & 0.475 & 1.526 & 0.945 & 0.704 & 0.620 & 0.572 & 0.556 & 1.663 & 1.081\\
    
    \textit{ETTm1$\rightarrow$ETTm2} & \textcolor{blue}{\underline{0.354}} & \textcolor{red}{\textbf{0.367}} & 0.405 & 0.440 & 0.446 & 0.447 & 0.469 & 0.484 & 0.424 & 0.463 & \textcolor{red}{\textbf{0.335}} & \textcolor{blue}{\underline{0.389}} & 1.521 & 0.951 & 0.603	& 0.578 & 0.466 & 0.495& 2.017 & 1.111\\
    
    \textit{ETTm2$\rightarrow$ETTh2} & 0.494 & 0.474 & 0.482 & 0.498 & 0.512 & 0.505 & \textcolor{red}{\textbf{0.423}} & \textcolor{red}{\textbf{0.439}} & 0.545 & 0.516 & \textcolor{blue}{\underline{0.455}} & \textcolor{blue}{\underline{0.471}} & 1.663 & 0.955 & 1.693	& 0.958 & 1.051 & 0.730 & 2.056 & 1.043\\
    
    \textit{ETTm2$\rightarrow$ETTm1} & \textcolor{red}{\textbf{0.588}} & \textcolor{red}{\textbf{0.487}} & 1.039 & 0.763 & 1.016 & 0.678 & 0.755 & 0.591 & 0.819 & 0.618 & \textcolor{blue}{\underline{0.649}} & \textcolor{blue}{\underline{0.537}} & 0.854 & 0.637 & 0.728	& 0.607 & 0.716 & 0.550& 0.941 & 0.698\\
\midrule
    1st Count & 4 & 5 & 0 & 0 & 0 & 0 & 2 & 2 & 0 & 0 & 1 & 0 & 0 & 0 & 0 & 0 & 1 & 1 & 0 & 0 \\
    \bottomrule
    \end{tabular}
}
  \label{tab:zero_shot_results}
\end{table*}

%% file: table/ablation_components.tex
\begin{table*}[!ht]
\centering
\caption{Ablation study results on different model components on the Weather dataset. We compare the full LDM4TS model with variants excluding key components: latent diffusion model (w/o LDM), vision encoder (w/o VE), temporal encoder (w/o TE), temporal conditioning (w/o TC), and frequency conditioning (w/o FC). \textit{\%Deg} denotes the performance degradation percentage compared to the full model.}
\scalebox{1}{
\begin{tabular}{@{}ccccccccccccc@{}}
\toprule
\multirow{2}{*}{Horizon} & \multicolumn{2}{c}{LDM4TS - Full} & \multicolumn{2}{c}{w/o LDM} & \multicolumn{2}{c}{w/o VE} & \multicolumn{2}{c}{w/o TE} & \multicolumn{2}{c}{w/o TC} & \multicolumn{2}{c}{w/o FC}\\
\cmidrule(lr){2-3} \cmidrule(lr){4-5} \cmidrule(lr){6-7} \cmidrule(lr){8-9} \cmidrule(lr){10-11} \cmidrule(lr){12-13}
& MSE & MAE & MSE & MAE & MSE & MAE & MSE & MAE & MSE & MAE & MSE & MAE \\
\midrule
96  & 0.162 & 0.213 & 0.164 & 0.216 & 0.213 & 0.266 & 0.213 & 0.266 & 0.162 & 0.214 & 0.162 & 0.214\\
192 & 0.219 & 0.267 & 0.224 & 0.274 & 0.259 & 0.298 & 0.259 & 0.299 & 0.206 & 0.257 & 0.207 & 0.257\\
336 & 0.260 & 0.297 & 0.280 & 0.311 & 0.267 & 0.302 & 0.276 & 0.311 & 0.260 & 0.295 & 0.260 & 0.295\\
720 & 0.338 & 0.350 & 0.364 & 0.364 & 0.342 & 0.357 & 0.337 & 0.354 & 0.336 & 0.348 & 0.354 & 0.370\\
\midrule
Avg & 0.245 & 0.282 & 0.258 & 0.291 & 0.270 & 0.306 & 0.271 & 0.307 & 0.241 & 0.278 & 0.246 & 0.284\\
\%Deg & -- & -- & \cellcolor{red!15}5.57\% & \cellcolor{red!15}3.35\%$\uparrow$ & \cellcolor{red!15}10.57\%$\uparrow$ & \cellcolor{red!15}8.57\%$\uparrow$ & \cellcolor{red!15}10.96\%$\uparrow$ & \cellcolor{red!15}9.12\%$\uparrow$ & \cellcolor{blue!15}-1.47\%$\downarrow$ & \cellcolor{blue!15}-1.16\%$\downarrow$ & \cellcolor{red!15}0.59\%$\uparrow$ & \cellcolor{red!15}0.90\%$\uparrow$ \\
\bottomrule
\end{tabular}
}
\label{tab:ablation_components}
\end{table*}

%% file: contents/05conclusion.tex
\section{Conclusion}
We present \model that adapts latent diffusion models with cross-modal conditioning mechanism for time series forecasting by transforming temporal data into multi-view visual representations. Our method significantly outperforms existing diffusion-based methods and specialized forecasting models, providing a novel vision-enhanced perspective to address the key challenges of intrinsic temporal pattern extraction and uncertainty modeling. Future work will focus on exploring diffusion models' potential in broader time series applications and developing comprehensive benchmarks for diffusion-based methods.

%% file: contents/06appendix.tex
\onecolumn
\newpage
\appendix

\input{appendix/A_Experimental_Details}
\input{appendix/B_Result_Details}
\input{appendix/D_Visualization}
\input{appendix/C_Algorithm}
\input{appendix/E_Pseudo}
\input{appendix/F_Interpretability_of_visual_transformation}
\input{appendix/G_limitation}
\newpage
\input{appendix/H_Efficiency}

%% file: appendix/A_Experimental_Details.tex
\section{Experimental Details}
\subsection{Dataset Details}
\label{appx:dataset_details} 
\input{table/dataset_detail}

We conduct experiments on the above real-world datasets to evaluate the performance of our proposed model and follow the same data processing and train-validation-test set split protocol used in TimesNet benchmark~\cite{wu2022timesnet}, ensuring a strict chronological order to prevent data leakage. Different datasets require specific adjustments to accommodate their unique characteristics:
\label{appx:dataset_configurations}

\paragraph{ETT Dataset~\cite{kim2021reversible}} The Electricity Transformer Temperature (ETT) dataset consists of both hourly (ETTh) and 15-minute (ETTm) frequency data, with 7 variables ($enc\_{in}$ = $dec\_{in}$ = $c\_{out}$ = 7) measuring transformer temperatures and related factors. For ETTh data, we set periodicity to 24 with hourly frequency, while ETTm data uses a periodicity of 96 with 15-minute intervals. Standard normalization is applied to each feature independently, and the model maintains the same architectural configuration across both temporal resolutions.

\paragraph{Traffic Dataset~\cite{wu2022timesnet}} The traffic flow dataset represents a high-dimensional scenario with 862 variables capturing traffic movements across different locations. To handle this scale, we implement gradient checkpointing and efficient attention mechanisms, complemented by progressive feature loading. The batch size is dynamically adjusted based on available GPU memory, and we maintain a periodicity of 24 to capture daily patterns. Our model employs specialized memory optimization techniques to process this large feature space efficiently.

\paragraph{ECL Dataset~\cite{wu2021autoformer}} The electricity consumption dataset contains 321 variables monitoring power usage patterns. We employ robust scaling techniques to handle outliers and implement sophisticated missing value imputation strategies. The model incorporates adaptive normalization layers to address the varying scales of electricity consumption across different regions and time periods. The daily periodicity is preserved through careful temporal encoding, while the high feature dimensionality is managed through efficient attention mechanisms.

\paragraph{Weather Dataset~\cite{wu2021autoformer}} This multivariate dataset encompasses 21 weather-related variables, each with distinct physical meanings and scale properties. Our approach implements feature-specific normalization to handle the diverse variable ranges while maintaining their physical relationships. The model captures both daily and seasonal patterns through enhanced temporal encoding, with special attention mechanisms designed to model the complex interactions between different weather variables. We maintain consistent prediction quality across all variables through carefully calibrated cross-attention mechanisms.

\subsection{Optimization Settings}
\label{appx:optimization_settings}

\subsubsection{Model Architecture Parameters}
\label{appx:model_parameters}
\input{table/model_parameters}
The core architecture of our diffusion-based model consists of several key components, each with specific parameter settings. The autoencoder pathway is configured with an image size of $64\times64$ and a patch size of $16$, providing an efficient latent representation while maintaining temporal information. The diffusion process uses $1000$ timesteps with carefully tuned noise scheduling ($\beta_{start} = 0.00085, \beta_{end} = 0.012$) to ensure stable training.

For the transformer backbone, we employ a configuration with $d_model = 256$ and $8$ attention heads, which empirically shows strong performance across different datasets. The encoder-decoder structure uses $2$ encoder layers and $1$ decoder layer, with a feed-forward dimension of $768$, striking a balance between model capacity and computational efficiency.

\subsubsection{Training Parameters}
\label{appx:training_settings}
\input{table/training_parameters}
We adopt a comprehensive training strategy with both general and task-specific parameters. The model is trained with a batch size of $32$ and an initial learning rate of $0.001$, using the \textit{AdamW} optimizer. Early stopping with a patience of $3$ epochs is implemented to prevent over-fitting. For time series processing, we use a sequence length of $96$ and a prediction length of $96$, with a label length of 48 for teacher forcing during training.

The training process employs automatic mixed precision (AMP) when available to accelerate training while maintaining numerical stability. We use MSE as the primary loss function, supplemented by additional regularization terms for specific tasks.

\subsection{Evaluation Metrics}
\label{appx:metric}
For evaluation metrics, we utilize the mean square error (MSE) and mean absolute error (MAE) for long-term forecasting. 
The calculations of these metrics are as follows:
\begin{align*} \label{equ:metrics}
    \text{MSE} &= \frac{1}{H}\sum_{h=1}^T (\mathbf{Y}_{h} - \Hat{\mathbf{Y}}_{h})^2,
    &
    \text{MAE} &= \frac{1}{H}\sum_{h=1}^H|\mathbf{Y}_{h} - \Hat{\mathbf{Y}}_{h}|,\\
\end{align*}
where $s$ is the periodicity of the time series data. $H$ denotes the number of data points (i.e., prediction horizon in our cases). $\mathbf{Y}_{h}$ and $\Hat{\mathbf{Y}}_{h}$ are the $h$-th ground truth and prediction where $h \in \{1, \cdots, H\}$.

\section{Details of Baseline Methods}
\label{appx:baselines}
We compare our approach with three categories of baseline methods used for comparative evaluation: transformer-based architectures, diffusion-based models, and other competitive approaches for time series forecasting.
\paragraph{Transformer-based Models:}
\textbf{FEDformer~\cite{zhou2022fedformer}} integrates wavelet decomposition with a Transformer architecture to efficiently capture multi-scale temporal dependencies by processing both time and frequency domains. 
\textbf{Autoformer~\cite{wu2021autoformer}} introduces a decomposing framework that separates the time series into trend and seasonal components, employing an autocorrelation mechanism for periodic pattern extraction.
\textbf{ETSformer~\cite{woo2022etsformer}} extends the classical exponential smoothing method with a Transformer architecture, decomposing time series into level, trend, and seasonal components while learning their interactions through attention mechanisms.
\textbf{Informer~\cite{zhou2021informer}} addresses the quadratic complexity issue of standard attention mechanisms through ProbSparse self-attention, which enables efficient handling of long input sequences.
\textbf{Reformer~\cite{kitaev2020reformer}} optimizes attention computation via Locality-Sensitive Hashing (LSH) and reversible residual networks, significantly reducing memory and computational costs.
\paragraph{Diffusion-based Models:}
\textbf{CSDI~\cite{tashiro2021csdi}} is tailored for irregularly-spaced time series, learning a score function of noise distribution under given conditions to generate samples for forecasting.
\textbf{ScoreGrad~\cite{song2020score}} utilizes a continuous-time framework for progressive denoising from Gaussian noise to reconstruct the original signal, allowing for adjustable step sizes during the denoising process.
\paragraph{Other Competitive Models:}
\textbf{DLinear~\cite{zeng2023transformers}} proposes a linear transformation approach directly on time series data, simplifying the prediction process under the assumption of linear changes over time.
\textbf{TimesNet~\cite{wu2022timesnet}} focuses on multi-scale feature extraction using various convolution kernels to capture temporal dependencies of different lengths, automatically selecting the most suitable feature scales.
\textbf{LightTS~\cite{campos2023lightts}} aims to build lightweight time series forecasting models, streamlining structures and parameters to reduce computational resource requirements while maintaining high predictive performance.

Each baseline method represents distinct paradigms within probabilistic generative modeling, attention-based architectures, and linear models, providing a comprehensive benchmark against which to evaluate LDM4TS.

%% file: table/dataset_detail.tex
\begin{table}[htbp]
  \caption{Summary of the benchmark datasets. Each dataset contains multiple time series (Dim.) with different sequence lengths, and is split into training, validation and testing sets. The data are collected at different frequencies across various domains.}
  \label{tab:dataset}
  \centering
  \scalebox{0.99}{
  \begin{tabular}{l|c|c|c|c|c}
    \toprule
Dataset & Dim. & Series Length & Dataset Size & Frequency & Domain \\
\toprule
ETTm1 & 7 & \{96, 192, 336, 720\} & (34465, 11521, 11521)  & 15 min & Temperature \\
ETTm2 & 7 & \{96, 192, 336, 720\} & (34465, 11521, 11521)  & 15 min & Temperature \\
ETTh1 & 7 & \{96, 192, 336, 720\} & (8545, 2881, 2881) & 1 hour & Temperature \\
ETTh2 & 7 & \{96, 192, 336, 720\} & (8545, 2881, 2881) & 1 hour & Temperature \\ 
Electricity & 321 & \{96, 192, 336, 720\} & (18317, 2633, 5261) & 1 hour & Electricity \\ 
Traffic & 862 & \{96, 192, 336, 720\} & (12185, 1757, 3509) & 1 hour & Transportation \\ 
Weather & 21 & \{96, 192, 336, 720\} & (36792, 5271, 10540) & 10 min 
& Weather \\

\bottomrule
\end{tabular}
}
\end{table}

%% file: table/model_parameters.tex
\begin{table}[htbp]
\centering
\caption{Default Model Architecture Parameters}
\begin{tabular}{|l|l|p{5.5cm}|}
\hline
\textbf{Parameter} & \textbf{Default Value} & \textbf{Description} \\
\hline
\multicolumn{3}{|l|}{\textit{Visual Representation Parameters}} \\
\hline
image\_size & 64 & Size of generated image representation \\
patch\_size & 16 & Size of patches for input processing \\
grayscale & True & Whether to use grayscale images \\
\hline
\multicolumn{3}{|l|}{\textit{Diffusion Process Parameters}} \\
\hline
num\_timesteps & 300 & Number of diffusion training steps \\
inference\_steps & 50 & Number of inference steps \\
beta\_start & 0.00085 & Initial value of noise schedule \\
beta\_end & 0.012 & Final value of noise schedule \\
use\_ddim & True & Whether to use DDIM sampler \\
unet\_layers & 1 & Number of layers in UNet \\
\hline
\multicolumn{3}{|l|}{\textit{Model Architecture Parameters}} \\
\hline
d\_model & 256 & Dimension of model hidden states \\
d\_ldm & 256 & Hidden dimension of LDM \\
d\_fusion & 256 & Dimension of gated fusion module \\
e\_layers & 2 & Number of encoder layers \\
d\_layers & 1 & Number of decoder layers \\
\hline
\multicolumn{3}{|l|}{\textit{Training Configuration}} \\
\hline
freeze\_ldm & True & Whether to freeze LDM parameters \\
save\_images & False & Whether to save generated images \\
output\_type & full & Type of output for ablation study \\
\hline
\end{tabular}
\end{table}

%% file: table/training_parameters.tex
\begin{table}[htbp]
\centering
\caption{Default Training Parameters}
\begin{tabular}{|l|l|p{5.5cm}|}
\hline
\textbf{Parameter} & \textbf{Default Value} & \textbf{Description} \\
\hline
\multicolumn{3}{|l|}{\textit{Basic Training Parameters}} \\
\hline
batch\_size & 32 & Number of samples per training batch \\
learning\_rate & 0.001 & Initial learning rate for optimization \\
train\_epochs & 10 & Total number of training epochs \\
patience & 3 & Epochs before early stopping \\
loss & MSE & Type of loss function \\
label\_len & 48 & Length of start token sequence \\
seq\_len & 96 & Length of input sequence \\
norm\_const & 0.4 & Coefficient for normalization \\
padding & 8 & Size of sequence padding \\
stride & 8 & Step size for sliding window \\
pred\_len & \makecell[l]{96/192/336/720} & Available prediction horizons \\
\hline
\multicolumn{3}{|l|}{\textit{Dataset-specific Parameters}} \\
\hline
c\_out & \makecell[l]{7 (ETTh1/h2/m1/m2) \\ 21 (Weather) \\ 321 (Electricity) \\ 862 (Traffic)} & Dataset-specific output dimensions \\
\hline
periodicity & \makecell[l]{24 (ETTh1/h2/Electricity/Traffic) \\ 96 (ETTm1/m2) \\ 144 (Weather)} & Natural cycle length per dataset \\
\hline
\end{tabular}
\end{table}

%% file: appendix/B_Result_Details.tex
\newpage
\subsection{Long-term Forecasting}
\label{appx:long_term_details}
\input{table/long_term_details}

\newpage
\subsection{Few-shot Forecasting}
\label{appx:few_shot_details}
\input{table/few_shot_details}
\input{table/few_shot_details_5p}

\subsection{Zero-shot Forecasting}
\label{appx:zero_shot_details}
\input{table/zero_shot_details}


%% file: table/long_term_details.tex
\begin{table}[!ht]
\centering
\caption{Details of long-term forecasting results.}
\vspace{-2mm}
\scalebox{0.78}{
\begin{tabular}{c|c|cc|cc|cc|cc|cc|cc|cc|cc|cc|cc}
\toprule
\multicolumn{2}{c|}{Methods} & \multicolumn{2}{c|}{LDM4TS} & \multicolumn{2}{c|}{CSDI} & \multicolumn{2}{c|}{ScoreGrad} & \multicolumn{2}{c|}{Autoformer} & \multicolumn{2}{c|}{FEDformer} & \multicolumn{2}{c|}{DLinear} & \multicolumn{2}{c|}{Informer} & \multicolumn{2}{c|}{TimesNet} & \multicolumn{2}{c|}{LightTS} & \multicolumn{2}{c}{Reformer} \\
\multicolumn{2}{c|}{Metric} & MSE & MAE & MSE & MAE & MSE & MAE & MSE & MAE & MSE & MAE & MSE & MAE & MSE & MAE & MSE & MAE & MSE & MAE & MSE & MAE \\ \midrule
\multirow{5}{*}{ETTh1} & 96 & 0.388 & 0.411 & 0.910 & 0.708 & 1.030 & 0.702 & 0.505 & 0.482 & 0.377 & 0.418 & 0.396 & 0.411 & 0.953 & 0.773 & 0.389 & 0.412 & 0.412 & 0.422 & 0.852 & 0.680 \\
& 192 & 0.412 & 0.430 & 0.975 & 0.738 & 1.033 & 0.705 & 0.475 & 0.470 & 0.420 & 0.444 & 0.445 & 0.440 & 1.016 & 0.790 & 0.439 & 0.442 & 0.581 & 0.545 & 1.008 & 0.742 \\
& 336 & 0.471 & 0.473 & 0.977 & 0.739 & 1.025 & 0.707 & 0.524 & 0.508 & 0.459 & 0.467 & 0.487 & 0.465 & 1.030 & 0.782 & 0.494 & 0.471 & 0.740 & 0.637 & 0.934 & 0.726 \\
& 720 & 0.501 & 0.502 & 1.086 & 0.783 & 1.036 & 0.721 & 0.512 & 0.509 & 0.502 & 0.503 & 0.513 & 0.510 & 1.161 & 0.853 & 0.517 & 0.494 & 0.625 & 0.574 & 1.229 & 0.831 \\
& \multicolumn{1}{c|}{Avg} & 0.443 & 0.454 & 0.987 & 0.742 & 1.031 & 0.709 & 0.504 & 0.492 & 0.439 & 0.458 & 0.460 & 0.457 & 1.040 & 0.799 & 0.460 & 0.455 & 0.590 & 0.544 & 1.006 & 0.745 \\ \midrule
\multirow{5}{*}{ETTh2} & 96 & 0.316 & 0.378 & 0.455 & 0.455 & 0.502 & 0.497 & 0.375 & 0.409 & 0.347 & 0.386 & 0.341 & 0.395 & 3.314 & 1.456 & 0.337 & 0.371 & 0.394 & 0.431 & 1.732 & 1.064 \\
& 192 & 0.356 & 0.404 & 0.493 & 0.506 & 0.509 & 0.501 & 0.463 & 0.461 & 0.422 & 0.437 & 0.482 & 0.479 & 6.750 & 2.137 & 0.405 & 0.415 & 0.635 & 0.550 & 2.662 & 1.305 \\
& 336 & 0.438 & 0.461 & 0.540 & 0.532 & 0.506 & 0.503 & 0.538 & 0.506 & 0.495 & 0.485 & 0.593 & 0.542 & 4.594 & 1.784 & 0.451 & 0.449 & 0.454 & 0.463 & 2.593 & 1.267 \\
& 720 & 0.436 & 0.465 & 0.668 & 0.588 & 0.531 & 0.520 & 0.491 & 0.498 & 0.493 & 0.494 & 0.840 & 0.661 & 3.545 & 1.592 & 0.435 & 0.449 & 3.556 & 1.267 & 3.135 & 1.339 \\
& \multicolumn{1}{c|}{Avg} & 0.387 & 0.427 & 0.539 & 0.520 & 0.512 & 0.505 & 0.467 & 0.468 & 0.439 & 0.451 & 0.564 & 0.519 & 4.551 & 1.742 & 0.407 & 0.421 & 1.260 & 0.678 & 2.531 & 1.244 \\ \midrule
\multirow{5}{*}{ETTm1} & 96 & 0.331 & 0.373 & 0.771 & 0.628 & 1.001 & 0.668 & 0.489 & 0.476 & 0.401 & 0.428 & 0.346 & 0.374 & 0.634 & 0.563 & 0.413 & 0.405 & 0.344 & 0.382 & 0.879 & 0.662 \\
& 192 & 0.346 & 0.382 & 0.789 & 0.640 & 1.008 & 0.673 & 0.599 & 0.532 & 0.465 & 0.462 & 0.382 & 0.391 & 0.729 & 0.617 & 0.481 & 0.436 & 0.348 & 0.381 & 0.963 & 0.736 \\
& 336 & 0.371 & 0.394 & 0.812 & 0.652 & 1.020 & 0.680 & 0.609 & 0.545 & 0.496 & 0.485 & 0.415 & 0.415 & 1.176 & 0.862 & 0.496 & 0.456 & 0.438 & 0.446 & 1.064 & 0.771 \\
& 720 & 0.362 & 0.397 & 0.852 & 0.675 & 1.033 & 0.691 & 0.608 & 0.551 & 0.521 & 0.504 & 0.473 & 0.451 & 0.929 & 0.717 & 0.518 & 0.475 & 0.576 & 0.539 & 1.147 & 0.777 \\
& \multicolumn{1}{c|}{Avg} & 0.352 & 0.387 & 0.806 & 0.649 & 1.016 & 0.678 & 0.576 & 0.526 & 0.471 & 0.470 & 0.404 & 0.408 & 0.867 & 0.690 & 0.477 & 0.443 & 0.427 & 0.437 & 1.013 & 0.737 \\ \midrule
\multirow{5}{*}{ETTm2} & 96 & 0.184 & 0.274 & 0.316 & 0.392 & 0.390 & 0.418 & 0.212 & 0.295 & 0.214 & 0.300 & 0.210 & 0.293 & 0.525 & 0.568 & 0.192 & 0.267 & 0.210 & 0.304 & 0.822 & 0.682 \\
& 192 & 0.334 & 0.382 & 0.349 & 0.408 & 0.418 & 0.433 & 0.279 & 0.338 & 0.286 & 0.355 & 0.269 & 0.328 & 1.051 & 0.761 & 0.253 & 0.304 & 0.419 & 0.459 & 1.465 & 0.916 \\
& 336 & 0.376 & 0.398 & 0.388 & 0.426 & 0.454 & 0.451 & 0.322 & 0.360 & 0.331 & 0.375 & 0.321 & 0.361 & 1.550 & 0.945 & 0.327 & 0.355 & 0.612 & 0.589 & 2.092 & 1.101 \\
& 720 & 0.436 & 0.465 & 0.473 & 0.476 & 0.521 & 0.486 & 0.414 & 0.411 & 0.439 & 0.435 & 0.415 & 0.413 & 3.247 & 1.359 & 0.422 & 0.405 & 2.077 & 1.102 & 3.116 & 1.335 \\
& \multicolumn{1}{c|}{Avg} & 0.333 & 0.380 & 0.381 & 0.425 & 0.446 & 0.447 & 0.307 & 0.351 & 0.318 & 0.366 & 0.304 & 0.349 & 1.593 & 0.908 & 0.299 & 0.333 & 0.830 & 0.614 & 1.874 & 1.009 \\ \midrule
\multirow{5}{*}{Weather} & 96 & 0.166 & 0.220 & 0.222 & 0.288 & 0.268 & 0.300 & 0.241 & 0.320 & 0.258 & 0.334 & 0.150 & 0.220 & 0.150 & 0.220 & 0.169 & 0.218 & 0.169 & 0.241 & 0.660 & 0.617 \\
& 192 & 0.214 & 0.262 & 0.267 & 0.319 & 0.317 & 0.334 & 0.305 & 0.363 & 0.303 & 0.356 & 0.203 & 0.270 & 0.203 & 0.270 & 0.227 & 0.265 & 0.243 & 0.304 & 0.951 & 0.768 \\
& 336 & 0.262 & 0.298 & 0.314 & 0.347 & 0.366 & 0.364 & 0.346 & 0.384 & 0.352 & 0.385 & 0.253 & 0.318 & 0.253 & 0.318 & 0.280 & 0.300 & 0.271 & 0.328 & 1.474 & 0.966 \\
& 720 & 0.341 & 0.352 & 0.381 & 0.387 & 0.434 & 0.405 & 0.423 & 0.434 & 0.419 & 0.427 & 0.374 & 0.415 & 0.374 & 0.415 & 0.386 & 0.371 & 0.355 & 0.386 & 1.832 & 1.082 \\
& \multicolumn{1}{c|}{Avg} & 0.245 & 0.283 & 0.296 & 0.335 & 0.347 & 0.351 & 0.329 & 0.375 & 0.333 & 0.375 & 0.245 & 0.306 & 0.245 & 0.306 & 0.265 & 0.288 & 0.259 & 0.315 & 1.229 & 0.858 \\ \midrule
\multirow{5}{*}{ECL} & 96 & 0.173 & 0.272 & 0.983 & 0.802 & 1.240 & 0.877 & 0.199 & 0.315 & 0.579 & 0.359 & 0.210 & 0.302 & 0.380 & 0.436 & 0.170 & 0.271 & 0.216 & 0.319 & 0.295 & 0.382 \\
& 192 & 0.182 & 0.283 & 0.997 & 0.805 & 1.248 & 0.880 & 0.276 & 0.358 & 0.619 & 0.381 & 0.210 & 0.305 & 0.352 & 0.426 & 0.193 & 0.288 & 0.227 & 0.329 & 0.334 & 0.412 \\
& 336 & 0.203 & 0.306 & 0.976 & 0.800 & 1.258 & 0.885 & 0.243 & 0.353 & 0.601 & 0.370 & 0.223 & 0.319 & 0.405 & 0.454 & 0.201 & 0.297 & 0.247 & 0.349 & 0.355 & 0.425 \\
& 720 & 0.236 & 0.334 & 1.015 & 0.810 & 1.286 & 0.895 & 0.293 & 0.380 & 0.651 & 0.399 & 0.258 & 0.350 & 0.378 & 0.436 & 0.266 & 0.355 & 0.283 & 0.376 & 0.319 & 0.396 \\
& \multicolumn{1}{c|}{Avg} & 0.199 & 0.299 & 0.993 & 0.804 & 1.258 & 0.884 & 0.253 & 0.352 & 0.612 & 0.377 & 0.225 & 0.319 & 0.378 & 0.438 & 0.208 & 0.303 & 0.243 & 0.343 & 0.326 & 0.404 \\ \midrule
\multirow{5}{*}{Traffic} & 96 & 0.529 & 0.315 & 1.697 & 0.859 & 2.077 & 1.016 & 0.334 & 0.528 & 0.579 & 0.359 & 0.649 & 0.396 & 0.774 & 0.432 & 0.600 & 0.316 & 0.619 & 0.401 & 0.736 & 0.418 \\
& 192 & 0.534 & 0.313 & 1.732 & 0.869 & 2.089 & 1.019 & 0.697 & 0.452 & 0.619 & 0.381 & 0.598 & 0.370 & 0.746 & 0.421 & 0.626 & 0.332 & 0.635 & 0.550 & 0.712 & 0.396 \\
& 336 & 0.541 & 0.317 & 1.753 & 0.873 & 2.108 & 1.023 & 0.754 & 0.492 & 0.601 & 0.370 & 0.605 & 0.373 & 0.826 & 0.463 & 0.638 & 0.340 & 0.454 & 0.463 & 0.712 & 0.393 \\
& 720 & 0.594 & 0.339 & 1.765 & 0.874 & 2.124 & 1.024 & 0.777 & 0.497 & 0.651 & 0.399 & 0.646 & 0.395 & 0.980 & 0.544 & 0.678 & 0.354 & 3.556 & 1.267 & 0.709 & 0.389 \\
& \multicolumn{1}{c|}{Avg} & 0.550 & 0.321 & 1.736 & 0.869 & 2.099 & 1.020 & 0.641 & 0.492 & 0.612 & 0.377 & 0.624 & 0.384 & 0.832 & 0.465 & 0.636 & 0.336 & 1.260 & 0.678 & 0.717 & 0.399 \\
\bottomrule
\end{tabular}
}
\end{table}

%% file: table/few_shot_details.tex
\begin{table}[!ht]
\centering
\caption{Details of few-shot forecasting results on 10\% training data.}
\vspace{-2mm}
\scalebox{0.78}{
\begin{tabular}{c|c|cc|cc|cc|cc|cc|cc|cc|cc|cc|cc}
\toprule
\multicolumn{2}{c|}{Methods} & \multicolumn{2}{c|}{LDM4TS} & \multicolumn{2}{c|}{CSDI} & \multicolumn{2}{c|}{ScoreGrad} & \multicolumn{2}{c|}{Autoformer} & \multicolumn{2}{c|}{FEDformer} & \multicolumn{2}{c|}{DLinear} & \multicolumn{2}{c|}{Informer} & \multicolumn{2}{c|}{TimesNet} & \multicolumn{2}{c|}{LightTS} & \multicolumn{2}{c}{Reformer} \\
\multicolumn{2}{c|}{Metric} & MSE & MAE & MSE & MAE & MSE & MAE & MSE & MAE & MSE & MAE & MSE & MAE & MSE & MAE & MSE & MAE & MSE & MAE & MSE & MAE \\ \midrule
\multirow{5}{*}{ETTh1} & 96 & 0.410 & 0.418 & 0.817 & 0.645 & 1.032 & 0.703 & 0.613 & 0.552 & 0.612 & 0.499 & 0.492 & 0.495 & 1.179 & 0.792 & 0.861 & 0.628 & 1.298 & 0.838 & 1.184 & 0.790 \\
& 192 & 0.443 & 0.443 & 0.835 & 0.653 & 1.033 & 0.705 & 0.722 & 0.598 & 0.624 & 0.555 & 0.565 & 0.538 & 1.199 & 0.806 & 0.797 & 0.593 & 1.322 & 0.854 & 1.295 & 0.850 \\
& 336 & 0.481 & 0.479 & 0.849 & 0.666 & 1.024 & 0.707 & 0.750 & 0.619 & 0.691 & 0.574 & 0.721 & 0.622 & 1.202 & 0.811 & 0.941 & 0.648 & 1.347 & 0.870 & 1.294 & 0.854 \\
& 720 & 0.549 & 0.534 & 0.895 & 0.698 & 1.036 & 0.721 & 0.721 & 0.616 & 0.728 & 0.614 & 0.986 & 0.743 & 1.217 & 0.825 & 0.877 & 0.641 & 1.534 & 0.947 & 1.223 & 0.838 \\
& \multicolumn{1}{c|}{Avg} & 0.471 & 0.468 & 0.849 & 0.665 & 1.031 & 0.709 & 0.702 & 0.596 & 0.639 & 0.561 & 0.691 & 0.600 & 1.199 & 0.809 & 0.869 & 0.628 & 1.375 & 0.877 & 1.249 & 0.833 \\ \midrule
\multirow{5}{*}{ETTh2} & 96 & 0.355 & 0.413 & 0.481 & 0.501 & 0.501 & 0.497 & 0.413 & 0.451 & 0.382 & 0.416 & 0.357 & 0.411 & 3.837 & 1.508 & 0.378 & 0.409 & 2.022 & 1.006 & 3.788 & 1.533 \\
& 192 & 0.406 & 0.435 & 0.535 & 0.527 & 0.509 & 0.502 & 0.474 & 0.477 & 0.478 & 0.474 & 0.569 & 0.519 & 3.856 & 1.513 & 0.430 & 0.467 & 2.329 & 1.104 & 3.552 & 1.483 \\
& 336 & 0.463 & 0.486 & 0.533 & 0.529 & 0.506 & 0.503 & 0.547 & 0.543 & 0.504 & 0.501 & 0.671 & 0.572 & 3.952 & 1.526 & 0.537 & 0.494 & 2.453 & 1.122 & 3.395 & 1.526 \\
& 720 & 0.583 & 0.505 & 0.558 & 0.537 & 0.531 & 0.520 & 0.516 & 0.523 & 0.499 & 0.509 & 0.824 & 0.648 & 3.842 & 1.503 & 0.510 & 0.491 & 3.816 & 1.407 & 3.205 & 1.401 \\
& \multicolumn{1}{c|}{Avg} & 0.452 & 0.460 & 0.527 & 0.523 & 0.512 & 0.505 & 0.488 & 0.499 & 0.466 & 0.475 & 0.605 & 0.538 & 3.872 & 1.513 & 0.479 & 0.465 & 2.655 & 1.160 & 3.485 & 1.486 \\ \midrule
\multirow{5}{*}{ETTm1} & 96 & 0.305 & 0.356 & 0.755 & 0.589 & 1.001 & 0.668 & 0.774 & 0.614 & 0.578 & 0.518 & 0.352 & 0.392 & 1.162 & 0.785 & 0.583 & 0.501 & 0.921 & 0.682 & 1.442 & 0.847 \\
& 192 & 0.345 & 0.378 & 0.771 & 0.596 & 1.008 & 0.673 & 0.754 & 0.592 & 0.617 & 0.556 & 0.382 & 0.412 & 1.172 & 0.793 & 0.630 & 0.528 & 0.957 & 0.701 & 1.444 & 0.862 \\
& 336 & 0.383 & 0.399 & 0.786 & 0.609 & 1.020 & 0.680 & 0.869 & 0.677 & 0.998 & 0.775 & 0.419 & 0.434 & 1.227 & 0.908 & 0.725 & 0.568 & 0.998 & 0.716 & 1.450 & 0.866 \\
& 720 & 0.452 & 0.439 & 0.825 & 0.629 & 1.033 & 0.691 & 0.810 & 0.630 & 0.693 & 0.579 & 0.490 & 0.477 & 1.207 & 0.797 & 0.769 & 0.549 & 1.007 & 0.719 & 1.366 & 0.850 \\
& \multicolumn{1}{c|}{Avg} & 0.371 & 0.393 & 0.784 & 0.606 & 1.015 & 0.678 & 0.802 & 0.628 & 0.722 & 0.635 & 0.411 & 0.429 & 1.192 & 0.821 & 0.677 & 0.537 & 0.971 & 0.705 & 1.426 & 0.856 \\ \midrule
\multirow{5}{*}{ETTm2} & 96 & 0.201 & 0.294 & 0.242 & 0.338 & 0.390 & 0.418 & 0.352 & 0.454 & 0.291 & 0.399 & 0.213 & 0.303 & 3.203 & 1.407 & 0.212 & 0.285 & 0.813 & 0.688 & 4.195 & 1.628 \\
& 192 & 0.288 & 0.342 & 0.296 & 0.364 & 0.418 & 0.433 & 0.694 & 0.691 & 0.307 & 0.379 & 0.278 & 0.345 & 3.112 & 1.387 & 0.270 & 0.323 & 1.008 & 0.768 & 4.042 & 1.601 \\
& 336 & 0.411 & 0.419 & 0.350 & 0.395 & 0.454 & 0.451 & 2.408 & 1.407 & 0.543 & 0.559 & 0.338 & 0.385 & 3.255 & 1.421 & 0.323 & 0.353 & 1.031 & 0.775 & 3.963 & 1.585 \\
& 720 & 0.443 & 0.435 & 0.447 & 0.446 & 0.521 & 0.488 & 1.913 & 1.166 & 0.712 & 0.614 & 0.436 & 0.440 & 3.909 & 1.543 & 0.474 & 0.449 & 1.096 & 0.791 & 3.711 & 1.532 \\
& \multicolumn{1}{c|}{Avg} & 0.336 & 0.373 & 0.334 & 0.385 & 0.446 & 0.447 & 1.342 & 0.930 & 0.463 & 0.488 & 0.316 & 0.368 & 3.370 & 1.440 & 0.320 & 0.353 & 0.987 & 0.756 & 3.978 & 1.587 \\ \midrule
\multirow{5}{*}{Weather} & 96 & 0.151 & 0.239 & 0.221 & 0.280 & 0.268 & 0.300 & 0.221 & 0.297 & 0.188 & 0.253 & 0.171 & 0.224 & 0.374 & 0.401 & 0.184 & 0.230 & 0.217 & 0.269 & 0.355 & 0.380 \\
& 192 & 0.187 & 0.247 & 0.265 & 0.317 & 0.317 & 0.334 & 0.270 & 0.322 & 0.250 & 0.304 & 0.215 & 0.263 & 0.552 & 0.478 & 0.245 & 0.283 & 0.259 & 0.304 & 0.522 & 0.462 \\
& 336 & 0.251 & 0.269 & 0.312 & 0.347 & 0.366 & 0.364 & 0.320 & 0.351 & 0.312 & 0.346 & 0.258 & 0.299 & 0.724 & 0.541 & 0.305 & 0.321 & 0.303 & 0.334 & 0.715 & 0.535 \\
& 720 & 0.328 & 0.349 & 0.381 & 0.387 & 0.434 & 0.405 & 0.390 & 0.396 & 0.387 & 0.393 & 0.320 & 0.346 & 0.739 & 0.558 & 0.381 & 0.371 & 0.377 & 0.382 & 0.511 & 0.500 \\
& \multicolumn{1}{c|}{Avg} & 0.229 & 0.276 & 0.295 & 0.333 & 0.347 & 0.351 & 0.300 & 0.342 & 0.284 & 0.324 & 0.241 & 0.283 & 0.597 & 0.495 & 0.279 & 0.301 & 0.289 & 0.322 & 0.526 & 0.469 \\ \midrule
\multirow{5}{*}{ECL} & 96 & 0.141 & 0.236 & 0.889 & 0.779 & 1.240 & 0.877 & 0.261 & 0.348 & 0.231 & 0.323 & 0.150 & 0.253 & 1.259 & 0.919 & 0.299 & 0.373 & 0.350 & 0.425 & 0.993 & 0.764 \\
& 192 & 0.158 & 0.260 & 0.896 & 0.781 & 1.248 & 0.880 & 0.338 & 0.406 & 0.261 & 0.356 & 0.164 & 0.264 & 1.160 & 0.873 & 0.305 & 0.379 & 0.376 & 0.448 & 0.998 & 0.795 \\
& 336 & 0.179 & 0.289 & 0.915 & 0.788 & 1.258 & 0.885 & 0.410 & 0.474 & 0.360 & 0.445 & 0.181 & 0.282 & 1.157 & 0.872 & 0.319 & 0.391 & 0.428 & 0.485 & 0.923 & 0.745 \\
& 720 & 0.209 & 0.314 & 0.936 & 0.792 & 1.286 & 0.895 & 0.715 & 0.685 & 0.530 & 0.585 & 0.223 & 0.321 & 1.203 & 0.898 & 0.369 & 0.426 & 0.611 & 0.597 & 1.004 & 0.790 \\
& \multicolumn{1}{c|}{Avg} & 0.172 & 0.275 & 0.909 & 0.785 & 1.258 & 0.884 & 0.431 & 0.478 & 0.346 & 0.427 & 0.180 & 0.280 & 1.195 & 0.891 & 0.323 & 0.392 & 0.441 & 0.489 & 0.980 & 0.769 \\ \midrule
\multirow{5}{*}{Traffic} & 96 & 0.615 & 0.360 & 1.701 & 0.868 & 2.078 & 1.016 & 0.672 & 0.405 & 0.639 & 0.416 & 0.942 & 0.571 & 1.557 & 0.821 & 0.719 & 0.416 & 1.157 & 0.836 & 1.527 & 0.815 \\
& 192 & 0.595 & 0.340 & 1.740 & 0.870 & 2.089 & 1.019 & 0.727 & 0.424 & 0.637 & 0.416 & 0.924 & 0.564 & 1.454 & 0.765 & 0.748 & 0.428 & 1.207 & 0.661 & 1.550 & 0.817 \\
& 336 & 0.611 & 0.352 & 1.746 & 0.871 & 2.107 & 1.023 & 0.749 & 0.454 & 0.655 & 0.427 & 0.941 & 0.569 & 1.521 & 0.812 & 0.853 & 0.471 & 1.334 & 0.713 & 1.550 & 0.819 \\
& 720 & 0.661 & 0.375 & 1.789 & 0.874 & 2.124 & 1.024 & 0.847 & 0.499 & 0.722 & 0.456 & 0.975 & 0.578 & 1.605 & 0.846 & 1.485 & 0.825 & 1.292 & 0.726 & 1.580 & 0.833 \\
& \multicolumn{1}{c|}{Avg} & 0.621 & 0.357 & 1.744 & 0.871 & 2.100 & 1.020 & 0.749 & 0.446 & 0.663 & 0.425 & 0.945 & 0.570 & 1.534 & 0.811 & 0.951 & 0.535 & 1.248 & 0.684 & 1.551 & 0.821 \\
\bottomrule
\end{tabular}
}
\label{tab:few_shot_detail}
\end{table}

%% file: table/few_shot_details_5p.tex
\begin{table}[!ht]
\centering
\caption{Details of few-shot results on 5\% training data. '-' means that 5\% data is not sufficient to constitute a training set.}
\vspace{-2mm}
\scalebox{0.78}{
\begin{tabular}{c|c|cc|cc|cc|cc|cc|cc|cc|cc|cc|cc}
\toprule
\multicolumn{2}{c|}{Methods} & \multicolumn{2}{c|}{LDM4TS} & \multicolumn{2}{c|}{CSDI} & \multicolumn{2}{c|}{ScoreGrad} & \multicolumn{2}{c|}{Autoformer} & \multicolumn{2}{c|}{FEDformer} & \multicolumn{2}{c|}{DLinear} & \multicolumn{2}{c|}{Informer} & \multicolumn{2}{c|}{TimesNet} & \multicolumn{2}{c|}{LightTS} & \multicolumn{2}{c}{Reformer} \\
\multicolumn{2}{c|}{Metric} & MSE & MAE & MSE & MAE & MSE & MAE & MSE & MAE & MSE & MAE & MSE & MAE & MSE & MAE & MSE & MAE & MSE & MAE & MSE & MAE \\ \midrule
\multirow{5}{*}{ETTh1} & 96 & 0.403 & 0.420 & 0.825 & 0.645 & 1.017 & 0.678 & 0.681 & 0.570 & 0.593 & 0.529 & 0.547 & 0.503 & 1.225 & 0.812 & 0.892 & 0.625 & 1.483 & 0.910 & 1.198 & 0.795 \\
& 192 & 0.464 & 0.458 & 0.842 & 0.659 & 1.032 & 0.690 & 0.725 & 0.602 & 0.652 & 0.563 & 0.720 & 0.604 & 1.249 & 0.828 & 0.940 & 0.665 & 1.525 & 0.930 & 1.273 & 0.853 \\
& 336 & 0.509 & 0.489 & 0.883 & 0.683 & 1.028 & 0.695 & 0.761 & 0.624 & 0.731 & 0.594 & 0.984 & 0.727 & 1.202 & 0.811 & 0.945 & 0.653 & 1.347 & 0.870 & 1.254 & 0.857 \\
& 720 & - & - & - & - & - & - & - & - & - & - & - & - & - & - & - & - & - & - & - & - \\
& \multicolumn{1}{c|}{Avg} & 0.458 & 0.456 & 0.850 & 0.662 & 1.026 & 0.687 & 0.722 & 0.599 & 0.659 & 0.562 & 0.750 & 0.611 & 1.225 & 0.817 & 0.926 & 0.648 & 1.452 & 0.903 & 1.242 & 0.835 \\
\midrule
\multirow{5}{*}{ETTh2} & 96 & 0.347 & 0.385 & 0.486 & 0.504 & 0.446 & 0.445 & 0.428 & 0.468 & 0.390 & 0.424 & 0.442 & 0.456 & 3.837 & 1.508 & 0.409 & 0.420 & 2.022 & 1.006 & 3.753 & 1.518 \\
& 192 & 0.538 & 0.510 & 0.521 & 0.520 & 0.520 & 0.483 & 0.496 & 0.504 & 0.457 & 0.465 & 0.617 & 0.542 & 3.975 & 1.933 & 0.483 & 0.464 & 3.534 & 1.348 & 3.516 & 1.473 \\
& 336 & 0.604 & 0.515 & 0.558 & 0.542 & 0.539 & 0.502 & 0.486 & 0.496 & 0.477 & 0.483 & 1.424 & 0.849 & 3.956 & 1.520 & 0.499 & 0.479 & 4.063 & 1.451 & 3.312 & 1.427 \\
& 720 & - & - & - & - & - & - & - & - & - & - & - & - & - & - & - & - & - & - & - & - \\
& \multicolumn{1}{c|}{Avg} & 0.496 & 0.470 & 0.522 & 0.522 & 0.502 & 0.477 & 0.470 & 0.489 & 0.441 & 0.457 & 0.828 & 0.616 & 3.923 & 1.654 & 0.464 & 0.454 & 3.206 & 1.268 & 3.527 & 1.473 \\
\midrule
\multirow{5}{*}{ETTm1} & 96 & 0.368 & 0.384 & 0.763 & 0.592 & 0.984 & 0.642 & 0.726 & 0.578 & 0.628 & 0.544 & 0.332 & 0.374 & 1.130 & 0.775 & 0.606 & 0.518 & 1.048 & 0.733 & 1.234 & 0.798 \\
& 192 & 0.384 & 0.400 & 0.772 & 0.594 & 1.003 & 0.653 & 0.750 & 0.591 & 0.666 & 0.566 & 0.358 & 0.390 & 1.150 & 0.788 & 0.681 & 0.539 & 1.097 & 0.756 & 1.287 & 0.839 \\
& 336 & 0.405 & 0.411 & 0.782 & 0.603 & 1.013 & 0.662 & 0.851 & 0.659 & 0.807 & 0.628 & 0.402 & 0.416 & 1.198 & 0.809 & 0.786 & 0.597 & 1.147 & 0.775 & 1.288 & 0.842 \\
& 720 & 0.471 & 0.452 & 0.819 & 0.623 & 1.038 & 0.679 & 0.857 & 0.655 & 0.822 & 0.633 & 0.511 & 0.489 & 1.175 & 0.794 & 0.796 & 0.593 & 1.200 & 0.799 & 1.247 & 0.828 \\
& \multicolumn{1}{c|}{Avg} & 0.407 & 0.412 & 0.784 & 0.603 & 1.009 & 0.659 & 0.796 & 0.621 & 0.731 & 0.593 & 0.401 & 0.417 & 1.163 & 0.792 & 0.717 & 0.562 & 1.123 & 0.766 & 1.264 & 0.827 \\
\midrule
\multirow{5}{*}{ETTm2} & 96 & 0.196 & 0.282 & 0.242 & 0.337 & 0.287 & 0.351 & 0.232 & 0.322 & 0.229 & 0.320 & 0.236 & 0.326 & 3.599 & 1.478 & 0.220 & 0.299 & 1.108 & 0.772 & 3.883 & 1.545 \\
& 192 & 0.277 & 0.331 & 0.298 & 0.367 & 0.342 & 0.381 & 0.291 & 0.357 & 0.394 & 0.381 & 0.306 & 0.373 & 3.578 & 1.475 & 0.311 & 0.361 & 1.317 & 0.850 & 3.553 & 1.484 \\
& 336 & 0.332 & 0.371 & 0.351 & 0.396 & 0.396 & 0.411 & 0.478 & 0.517 & 0.378 & 0.427 & 0.380 & 0.423 & 3.561 & 1.473 & 0.338 & 0.366 & 1.415 & 0.879 & 3.446 & 1.460 \\
& 720 & 0.441 & 0.427 & 0.451 & 0.448 & 0.491 & 0.458 & 0.553 & 0.538 & 0.523 & 0.510 & 0.674 & 0.583 & 3.896 & 1.533 & 0.509 & 0.465 & 1.822 & 0.984 & 3.445 & 1.460 \\
& \multicolumn{1}{c|}{Avg} & 0.311 & 0.353 & 0.335 & 0.387 & 0.379 & 0.400 & 0.388 & 0.433 & 0.381 & 0.405 & 0.399 & 0.426 & 3.659 & 1.490 & 0.345 & 0.373 & 1.416 & 0.871 & 3.582 & 1.487 \\
\midrule
\multirow{5}{*}{Weather} & 96 & 0.178 & 0.230 & 0.221 & 0.280 & 0.268 & 0.300 & 0.227 & 0.299 & 0.229 & 0.309 & 0.184 & 0.242 & 0.497 & 0.497 & 0.207 & 0.253 & 0.230 & 0.285 & 0.406 & 0.435 \\
& 192 & 0.223 & 0.270 & 0.265 & 0.317 & 0.317 & 0.334 & 0.278 & 0.333 & 0.265 & 0.317 & 0.228 & 0.283 & 0.620 & 0.545 & 0.272 & 0.307 & 0.274 & 0.323 & 0.446 & 0.450 \\
& 336 & 0.280 & 0.314 & 0.312 & 0.347 & 0.366 & 0.364 & 0.351 & 0.393 & 0.353 & 0.392 & 0.279 & 0.322 & 0.649 & 0.547 & 0.313 & 0.328 & 0.318 & 0.355 & 0.465 & 0.459 \\
& 720 & 0.350 & 0.363 & 0.381 & 0.387 & 0.434 & 0.405 & 0.387 & 0.389 & 0.391 & 0.394 & 0.364 & 0.388 & 0.570 & 0.522 & 0.400 & 0.385 & 0.401 & 0.418 & 0.471 & 0.468 \\
& \multicolumn{1}{c|}{Avg} & 0.258 & 0.294 & 0.295 & 0.333 & 0.347 & 0.351 & 0.311 & 0.354 & 0.310 & 0.353 & 0.264 & 0.309 & 0.584 & 0.528 & 0.298 & 0.318 & 0.306 & 0.345 & 0.447 & 0.453 \\
\midrule
\multirow{5}{*}{ECL} & 96 & 0.190 & 0.283 & 0.893 & 0.777 & 1.231 & 0.871 & 0.297 & 0.367 & 0.235 & 0.322 & 0.231 & 0.325 & 1.265 & 0.919 & 0.315 & 0.389 & 0.639 & 0.609 & 1.414 & 0.855 \\
& 192 & 0.193 & 0.287 & 0.906 & 0.782 & 1.233 & 0.872 & 0.308 & 0.375 & 0.247 & 0.341 & 0.231 & 0.329 & 1.298 & 0.939 & 0.318 & 0.396 & 0.772 & 0.678 & 1.240 & 0.919 \\
& 336 & 0.208 & 0.303 & 0.919 & 0.788 & 1.244 & 0.876 & 0.354 & 0.411 & 0.267 & 0.356 & 0.243 & 0.342 & 1.302 & 0.942 & 0.340 & 0.415 & 0.901 & 0.745 & 1.253 & 0.921 \\
& 720 & 0.242 & 0.333 & 0.947 & 0.796 & 1.274 & 0.888 & 0.426 & 0.466 & 0.318 & 0.394 & 0.277 & 0.370 & 1.269 & 0.919 & 0.635 & 0.613 & 1.200 & 0.871 & 1.249 & 0.921 \\
& \multicolumn{1}{c|}{Avg} & 0.208 & 0.301 & 0.916 & 0.786 & 1.245 & 0.877 & 0.346 & 0.405 & 0.267 & 0.353 & 0.246 & 0.342 & 1.281 & 0.930 & 0.402 & 0.453 & 0.878 & 0.726 & 1.289 & 0.904 \\
\midrule
\multirow{5}{*}{Traffic} & 96 & 0.642 & 0.376 & 1.701 & 0.868 & 2.078 & 1.016 & 0.795 & 0.481 & 0.670 & 0.421 & 0.748 & 0.459 & 1.557 & 0.821 & 0.854 & 0.492 & 1.157 & 0.636 & 1.586 & 0.841 \\
& 192 & 0.626 & 0.360 & 1.740 & 0.870 & 2.089 & 1.019 & 0.837 & 0.503 & 0.653 & 0.405 & 0.705 & 0.442 & 1.596 & 0.834 & 0.894 & 0.517 & 1.688 & 0.848 & 1.602 & 0.844 \\
& 336 & 0.627 & 0.362 & 1.746 & 0.871 & 2.107 & 1.023 & 0.867 & 0.523 & 0.707 & 0.445 & 0.635 & 0.397 & 1.621 & 0.841 & 0.853 & 0.471 & 1.826 & 0.903 & 1.668 & 0.868 \\
& 720 & - & - & - & - & - & - & - & - & - & - & - & - & - & - & - & - & - & - & - & - \\
& \multicolumn{1}{c|}{Avg} & 0.632 & 0.366 & 1.729 & 0.870 & 2.092 & 1.019 & 0.833 & 0.502 & 0.677 & 0.424 & 0.696 & 0.433 & 1.591 & 0.832 & 0.867 & 0.493 & 1.557 & 0.796 & 1.619 & 0.851 \\
\bottomrule
\end{tabular}
}
\label{tab:few_shot_5p_detail}
\end{table}

%% file: table/zero_shot_details.tex
\begin{table}[!ht]
\centering
\caption{Details of Zero-shot forecasting results.}
\vspace{-2mm}
\scalebox{0.78}{
\begin{tabular}{c|c|cc|cc|cc|cc|cc|cc|cc|cc|cc|cc}
\toprule
\multicolumn{2}{c|}{Methods} & \multicolumn{2}{c|}{LDM4TS} & \multicolumn{2}{c|}{CSDI} & \multicolumn{2}{c|}{ScoreGrad} & \multicolumn{2}{c|}{Autoformer} & \multicolumn{2}{c|}{FEDformer} & \multicolumn{2}{c|}{DLinear} & \multicolumn{2}{c|}{Informer} & \multicolumn{2}{c|}{ETSformer} & \multicolumn{2}{c|}{LightTS} & \multicolumn{2}{c}{Reformer} \\
\multicolumn{2}{c|}{Metric} & MSE & MAE & MSE & MAE & MSE & MAE & MSE & MAE & MSE & MAE & MSE & MAE & MSE & MAE & MSE & MAE & MSE & MAE & MSE & MAE \\ \midrule
\multirow{5}{*}{\makecell{ETTh1$\rightarrow$\\ETTh2}} & 96 & 0.349 & 0.383 & 0.463 & 0.489 & 0.502 & 0.497 & 0.469 & 0.486 & 0.414 & 0.447 & 0.347 & 0.400 & 1.834 & 1.058 & 0.381 & 0.427 & 0.376 & 0.424 & 1.833 & 1.060 \\
& 192 & 0.435 & 0.434 & 0.483 & 0.501 & 0.509 & 0.502 & 0.456 & 0.567 & 0.514 & 0.508 & 0.447 & 0.460 & 1.738 & 1.038 & 0.427 & 0.603 & 0.681 & 0.577 & 1.872 & 1.080 \\
& 336 & 0.478 & 0.465 & 0.477 & 0.559 & 0.506 & 0.503 & 0.655 & 0.588 & 0.533 & 0.525 & 0.515 & 0.505 & 2.513 & 1.212 & 0.658 & 0.615 & 1.169 & 0.795 & 2.004 & 1.114 \\
& 720 & 0.572 & 0.527 & 0.577 & 0.559 & 0.531 & 0.520 & 0.570 & 0.549 & 0.520 & 0.522 & 0.665 & 0.589 & 3.082 & 1.369 & 0.889 & 0.713 & 2.075 & 1.000 & 2.766 & 1.245 \\
& \multicolumn{1}{c|}{Avg} & 0.458 & 0.452 & 0.500 & 0.527 & 0.512 & 0.505 & 0.582 & 0.548 & 0.495 & 0.501 & 0.493 & 0.488 & 2.292 & 1.169 & 0.589 & 0.589 & 1.075 & 0.699 & 2.119 & 1.125 \\ \midrule
\multirow{5}{*}{\makecell{ETTh1 $\rightarrow$\\ ETTm2}} & 96 & 0.227 & 0.316 & 0.343 & 0.410 & 0.390 & 0.418 & 0.352 & 0.432 & 0.277 & 0.367 & 0.255 & 0.357 & 1.803 & 1.055 & 0.391 & 0.482 & 0.307 & 0.402 & 1.963 & 1.107 \\
& 192 & 0.312 & 0.373 & 0.383 & 0.430 & 0.418 & 0.433 & 0.413 & 0.460 & 0.329 & 0.399 & 0.338 & 0.413 & 1.682 & 1.017 & 0.514 & 0.514 & 0.614 & 0.560 & 2.007 & 1.131 \\
& 336 & 0.368 & 0.407 & 0.417 & 0.444 & 0.454 & 0.451 & 0.465 & 0.489 & 0.400 & 0.447 & 0.425 & 0.465 & 2.228 & 1.128 & 0.514 & 0.575 & 1.184 & 0.812 & 2.121 & 1.151 \\
& 720 & 0.569 & 0.505 & 0.499 & 0.489 & 0.521 & 0.486 & 0.599 & 0.551 & 0.485 & 0.482 & 0.640 & 0.573 & 2.954 & 1.297 & 0.857 & 0.699 & 2.128 & 1.028 & 2.822 & 1.272 \\
& \multicolumn{1}{c|}{Avg} & 0.369 & 0.400 & 0.410 & 0.444 & 0.446 & 0.447 & 0.457 & 0.483 & 0.373 & 0.424 & 0.415 & 0.452 & 2.167 & 1.124 & 0.569 & 0.568 & 1.058 & 0.700 & 2.228 & 1.165 \\ \midrule
\multirow{5}{*}{\makecell{ETTh2 $\rightarrow$\\ ETTh1}} & 96 & 0.706 & 0.540 & 1.318 & 0.879 & 1.033 & 0.703 & 0.693 & 0.569 & 2.310 & 1.238 & 0.689 & 0.555 & 1.157 & 0.781 & 0.863 & 0.651 & 0.528 & 0.486 & 0.902 & 0.689 \\
& 192 & 0.691 & 0.564 & 1.361 & 0.891 & 1.034 & 0.705 & 0.760 & 0.601 & 0.793 & 0.622 & 0.707 & 0.568 & 1.673 & 0.995 & 0.822 & 0.663 & 0.554 & 0.503 & 0.903 & 0.717 \\
& 336 & 0.697 & 0.576 & 1.431 & 0.901 & 1.025 & 0.707 & 0.781 & 0.619 & 0.883 & 0.662 & 0.710 & 0.577 & 1.730 & 0.978 & 0.883 & 0.696 & 0.598 & 0.535 & 0.933 & 0.713 \\
& 720 & 0.796 & 0.630 & 1.555 & 0.920 & 1.037 & 0.721 & 0.796 & 0.644 & 0.906 & 0.692 & 0.704 & 0.596 & 2.291 & 1.158 & 0.886 & 0.690 & 0.586 & 0.547 & 0.895 & 0.700 \\
& \multicolumn{1}{c|}{Avg} & 0.723 & 0.577 & 1.416 & 0.898 & 1.032 & 0.709 & 0.757 & 0.608 & 1.423 & 0.803 & 0.703 & 0.574 & 1.713 & 0.978 & 0.864 & 0.675 & 0.567 & 0.518 & 0.909 & 0.705 \\ \midrule
\multirow{5}{*}{\makecell{ETTh2 $\rightarrow$\\ ETTm2}} & 96 & 0.286 & 0.373 & 0.323 & 0.397 & 0.390 & 0.418 & 0.263 & 0.352 & 0.480 & 0.518 & 0.240 & 0.336 & 2.789 & 1.353 & 0.646 & 0.622 & 0.285 & 0.385 & 2.115 & 1.141 \\
& 192 & 0.326 & 0.391 & 0.363 & 0.418 & 0.418 & 0.433 & 0.326 & 0.389 & 0.328 & 0.383 & 0.295 & 0.369 & 8.210 & 2.073 & 1.370 & 0.965 & 0.482 & 0.499 & 2.830 & 1.282 \\
& 336 & 0.473 & 0.461 & 0.399 & 0.434 & 0.454 & 0.451 & 0.387 & 0.426 & 0.427 & 0.459 & 0.345 & 0.397 & 5.368 & 1.877 & 1.893 & 1.158 & 0.799 & 0.656 & 2.702 & 1.242 \\
& 720 & 0.644 & 0.551 & 0.502 & 0.497 & 0.521 & 0.486 & 0.467 & 0.478 & 0.496 & 0.481 & 0.432 & 0.442 & 4.054 & 1.660 & 1.371 & 0.921 & 1.246 & 0.801 & 3.257 & 1.363 \\
& \multicolumn{1}{c|}{Avg} & 0.432 & 0.444 & 0.397 & 0.437 & 0.446 & 0.447 & 0.386 & 0.411 & 0.433 & 0.460 & 0.328 & 0.386 & 4.606 & 1.763 & 1.320 & 0.917 & 0.703 & 0.585 & 2.726 & 1.257 \\ \midrule
\multirow{5}{*}{\makecell{ETTm1 $\rightarrow$\\ ETTh2}} & 96 & 0.390 & 0.402 & 0.466 & 0.492 & 0.502 & 0.497 & 0.435 & 0.470 & 0.573 & 0.556 & 0.365 & 0.415 & 1.377 & 0.894 & 0.474 & 0.500 & 0.388 & 0.444 & 1.621 & 1.008 \\
& 192 & 0.471 & 0.439 & 0.484 & 0.504 & 0.508 & 0.501 & 0.495 & 0.489 & 0.582 & 0.551 & 0.454 & 0.462 & 1.715 & 1.011 & 0.778 & 0.666 & 0.517 & 0.523 & 2.130 & 1.177 \\
& 336 & 0.496 & 0.460 & 0.506 & 0.514 & 0.506 & 0.503 & 0.470 & 0.472 & 0.578 & 0.563 & 0.496 & 0.494 & 1.519 & 0.940 & 0.668 & 0.601 & 0.576 & 0.575 & 1.844 & 1.056 \\
& 720 & 0.450 & 0.436 & 0.561 & 0.549 & 0.503 & 0.520 & 0.480 & 0.485 & 0.619 & 0.582 & 0.541 & 0.529 & 1.492 & 0.936 & 0.894 & 0.712 & 0.797 & 0.682 & 1.056 & 1.081 \\
& \multicolumn{1}{c|}{Avg} & 0.452 & 0.434 & 0.504 & 0.515 & 0.505 & 0.505 & 0.470 & 0.479 & 0.587 & 0.565 & 0.464 & 0.475 & 1.526 & 0.945 & 0.704 & 0.620 & 0.572 & 0.556 & 1.663 & 1.081 \\ \midrule
\multirow{5}{*}{\makecell{ETTm1 $\rightarrow$\\ ETTm2}} & 96 & 0.220 & 0.287 & 0.341 & 0.409 & 0.390 & 0.418 & 0.385 & 0.457 & 0.335 & 0.425 & 0.221 & 0.314 & 1.333 & 0.883 & 0.348 & 0.433 & 0.260 & 0.357 & 1.568 & 0.991 \\
& 192 & 0.284 & 0.335 & 0.374 & 0.424 & 0.418 & 0.433 & 0.433 & 0.469 & 0.384 & 0.439 & 0.286 & 0.359 & 1.617 & 0.985 & 0.651 & 0.625 & 0.366 & 0.442 & 2.271 & 1.223 \\
& 336 & 0.385 & 0.391 & 0.418 & 0.445 & 0.454 & 0.451 & 0.476 & 0.477 & 0.434 & 0.472 & 0.357 & 0.406 & 1.519 & 0.953 & 0.551 & 0.551 & 0.467 & 0.517 & 1.964 & 1.090 \\
& 720 & 0.574 & 0.484 & 0.485 & 0.484 & 0.521 & 0.486 & 0.582 & 0.535 & 0.543 & 0.516 & 0.476 & 0.476 & 1.615 & 0.985 & 0.863 & 0.701 & 0.772 & 0.666 & 2.264 & 1.142 \\
& \multicolumn{1}{c|}{Avg} & 0.366 & 0.374 & 0.405 & 0.440 & 0.446 & 0.447 & 0.469 & 0.484 & 0.424 & 0.463 & 0.335 & 0.389 & 1.521 & 0.951 & 0.603 & 0.578 & 0.466 & 0.495 & 2.017 & 1.111 \\ \midrule
\multirow{5}{*}{\makecell{ETTm2 $\rightarrow$\\ ETTh2}} & 96 & 0.368 & 0.416 & 0.425 & 0.467 & 0.502 & 0.497 & 0.353 & 0.393 & 0.773 & 0.637 & 0.333 & 0.391 & 1.877 & 0.699 & 0.608 & 0.608 & 0.381 & 0.433 & 1.006 & 0.750 \\
& 192 & 0.503 & 0.475 & 0.454 & 0.485 & 0.509 & 0.502 & 0.432 & 0.437 & 0.455 & 0.457 & 0.441 & 0.458 & 1.364 & 0.912 & 0.551 & 0.567 & 0.620 & 0.589 & 1.561 & 0.914 \\
& 336 & 0.595 & 0.529 & 0.475 & 0.493 & 0.507 & 0.503 & 0.452 & 0.459 & 0.482 & 0.485 & 0.505 & 0.503 & 1.449 & 0.929 & 0.886 & 0.770 & 0.911 & 0.733 & 2.243 & 1.121 \\
& 720 & 0.653 & 0.556 & 0.576 & 0.547 & 0.531 & 0.520 & 0.453 & 0.467 & 0.470 & 0.486 & 0.543 & 0.534 & 2.922 & 1.282 & 4.727 & 1.887 & 2.292 & 1.164 & 3.413 & 1.385 \\
& \multicolumn{1}{c|}{Avg} & 0.535 & 0.494 & 0.482 & 0.498 & 0.512 & 0.505 & 0.423 & 0.439 & 0.545 & 0.516 & 0.455 & 0.471 & 1.663 & 0.955 & 1.693 & 0.958 & 1.051 & 0.730 & 2.056 & 1.043 \\ \midrule
\multirow{5}{*}{\makecell{ETTm2 $\rightarrow$\\ ETTm1}} & 96 & 0.488 & 0.431 & 0.986 & 0.743 & 1.000 & 0.668 & 0.735 & 0.576 & 0.761 & 0.595 & 0.570 & 0.490 & 0.899 & 0.623 & 0.736 & 0.601 & 0.835 & 0.551 & 0.848 & 0.637 \\
& 192 & 0.591 & 0.497 & 1.057 & 0.764 & 1.009 & 0.673 & 0.753 & 0.586 & 0.857 & 0.632 & 0.590 & 0.506 & 0.816 & 0.621 & 0.703 & 0.591 & 0.773 & 0.554 & 0.915 & 0.687 \\
& 336 & 0.640 & 0.503 & 1.004 & 0.754 & 1.020 & 0.680 & 0.750 & 0.593 & 0.779 & 0.600 & 0.706 & 0.567 & 0.823 & 0.634 & 0.731 & 0.613 & 0.592 & 0.508 & 1.090 & 0.777 \\
& 720 & 0.631 & 0.515 & 1.107 & 0.791 & 1.033 & 0.691 & 0.782 & 0.609 & 0.879 & 0.645 & 0.731 & 0.584 & 0.877 & 0.672 & 0.740 & 0.624 & 0.665 & 0.587 & 0.910 & 0.690 \\
& \multicolumn{1}{c|}{Avg} & 0.588 & 0.487 & 1.039 & 0.763 & 1.016 & 0.678 & 0.755 & 0.591 & 0.819 & 0.618 & 0.649 & 0.537 & 0.854 & 0.637 & 0.728 & 0.607 & 0.716 & 0.550 & 0.941 & 0.698 \\
\bottomrule
\end{tabular}
}
\end{table}

%% file: appendix/D_Visualization.tex
\newpage
\section{Interpretability of LDM4TS}
\subsection{Showcases}
\label{appx:showcases}
\begin{figure}[!ht]
  \centering
  \includegraphics[width=\linewidth]{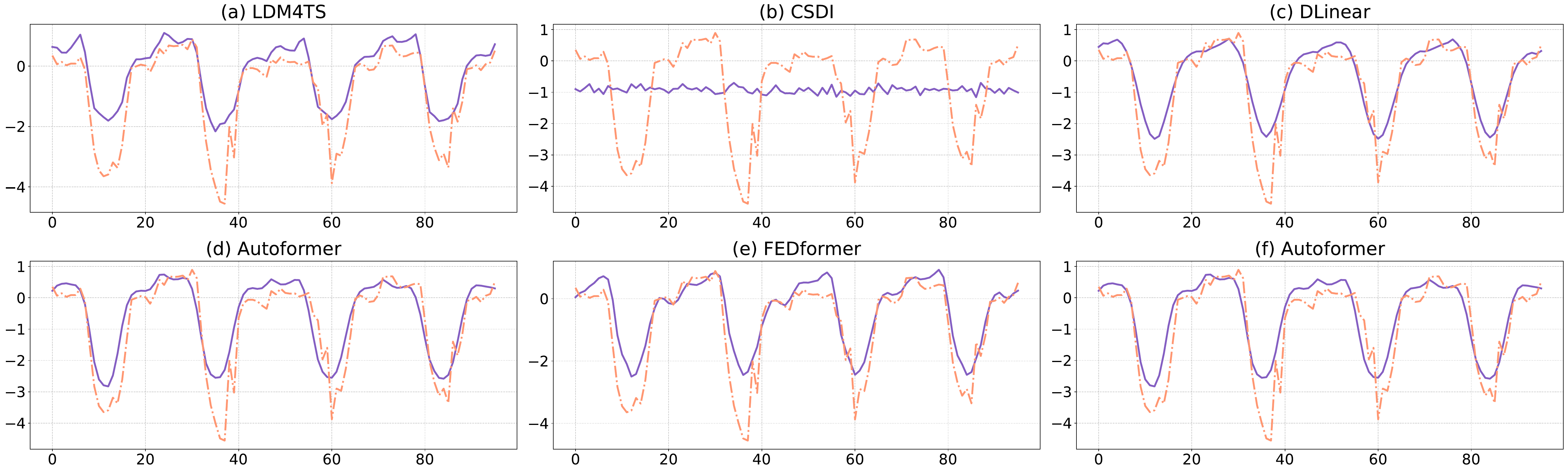}
\caption{Visualization of ETTh1 predictions by different models under the input-96-predict-96 setting. The orange lines stand for the ground truth and the blue lines stand for predicted values.}
\label{fig:showcases_96}
\end{figure}

\begin{figure}[!ht]
  \centering
  \includegraphics[width=\linewidth]{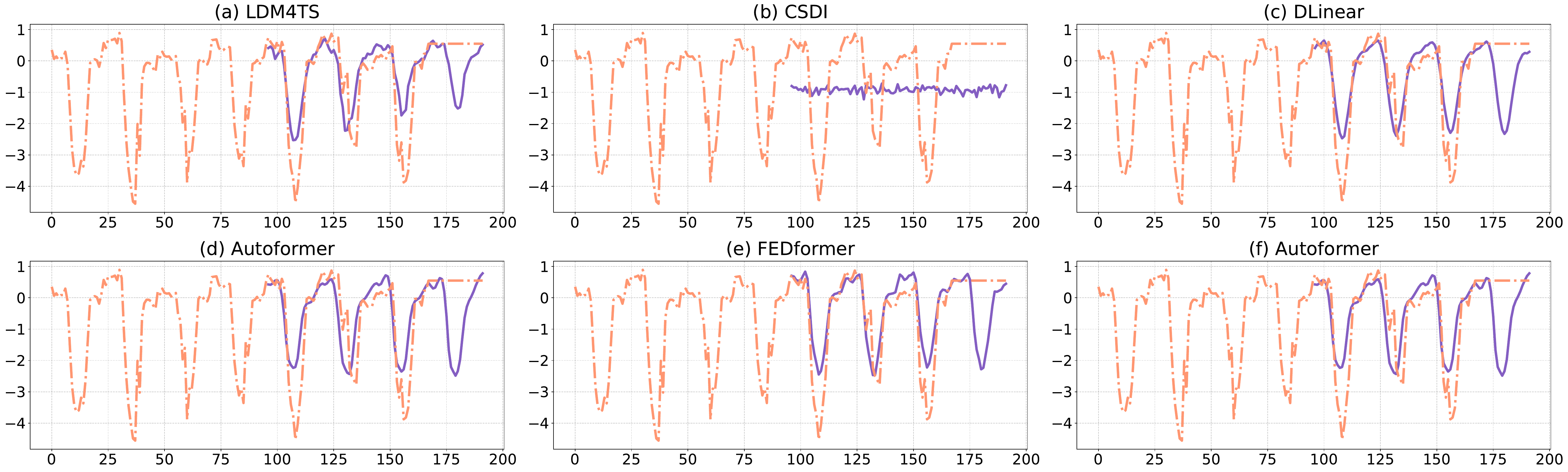}
\caption{Visualization of ETTh1 predictions by different models under the input-96-predict-192 setting. The orange lines stand for the ground truth and the blue lines stand for predicted values.}
\label{fig:showcases_192}
\end{figure}

\begin{figure}[!ht]
  \centering
  \includegraphics[width=\linewidth]{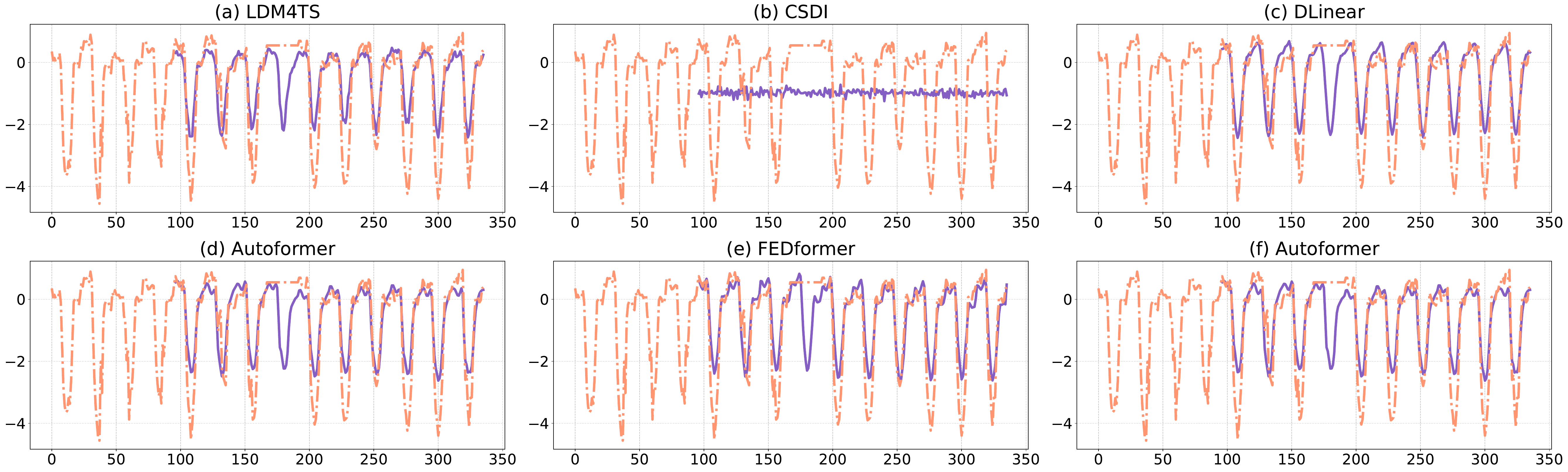}
\caption{Visualization of ETTh1 predictions by different models under the input-96-predict-336 setting. The orange lines stand for the ground truth and the blue lines stand for predicted values.}
\label{fig:showcases_336}
\end{figure}

\begin{figure}[!ht]
  \centering
  \includegraphics[width=\linewidth]{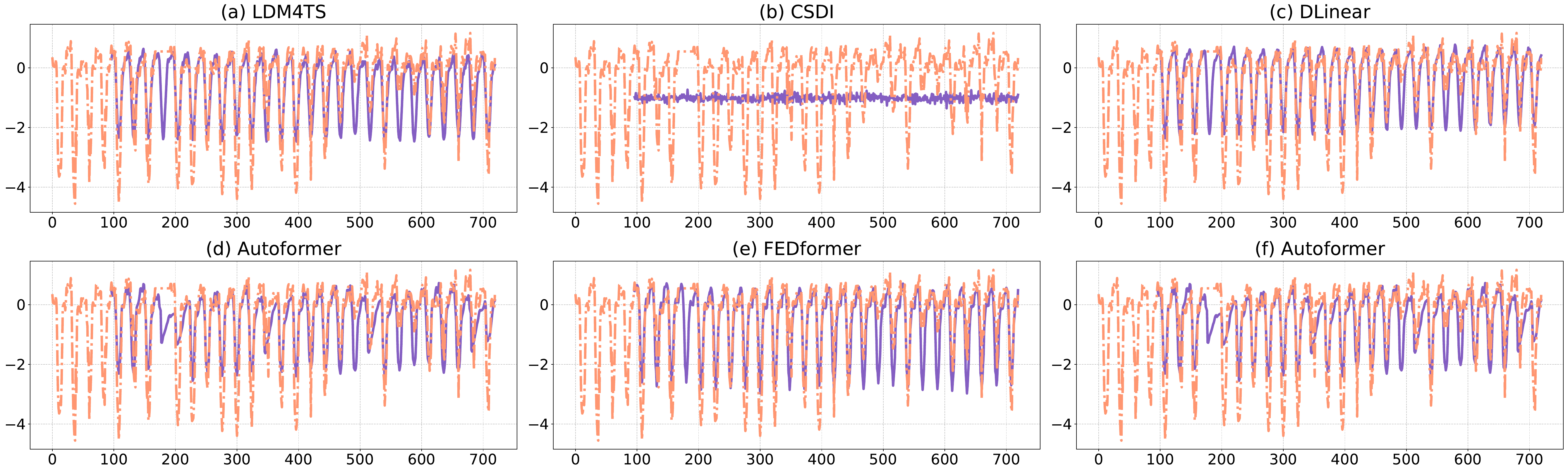}
\caption{Visualization of ETTh1 predictions by different models under the input-96-predict-720 setting. The orange lines stand for the ground truth and the blue lines stand for predicted values.}
\label{fig:showcases_720}
\end{figure}

\newpage
\subsection{Visualization of Pixel Space}
\label{appx:visualization_picel_space}
\begin{figure*}[!ht]
  \centering
  \includegraphics[width=\linewidth]{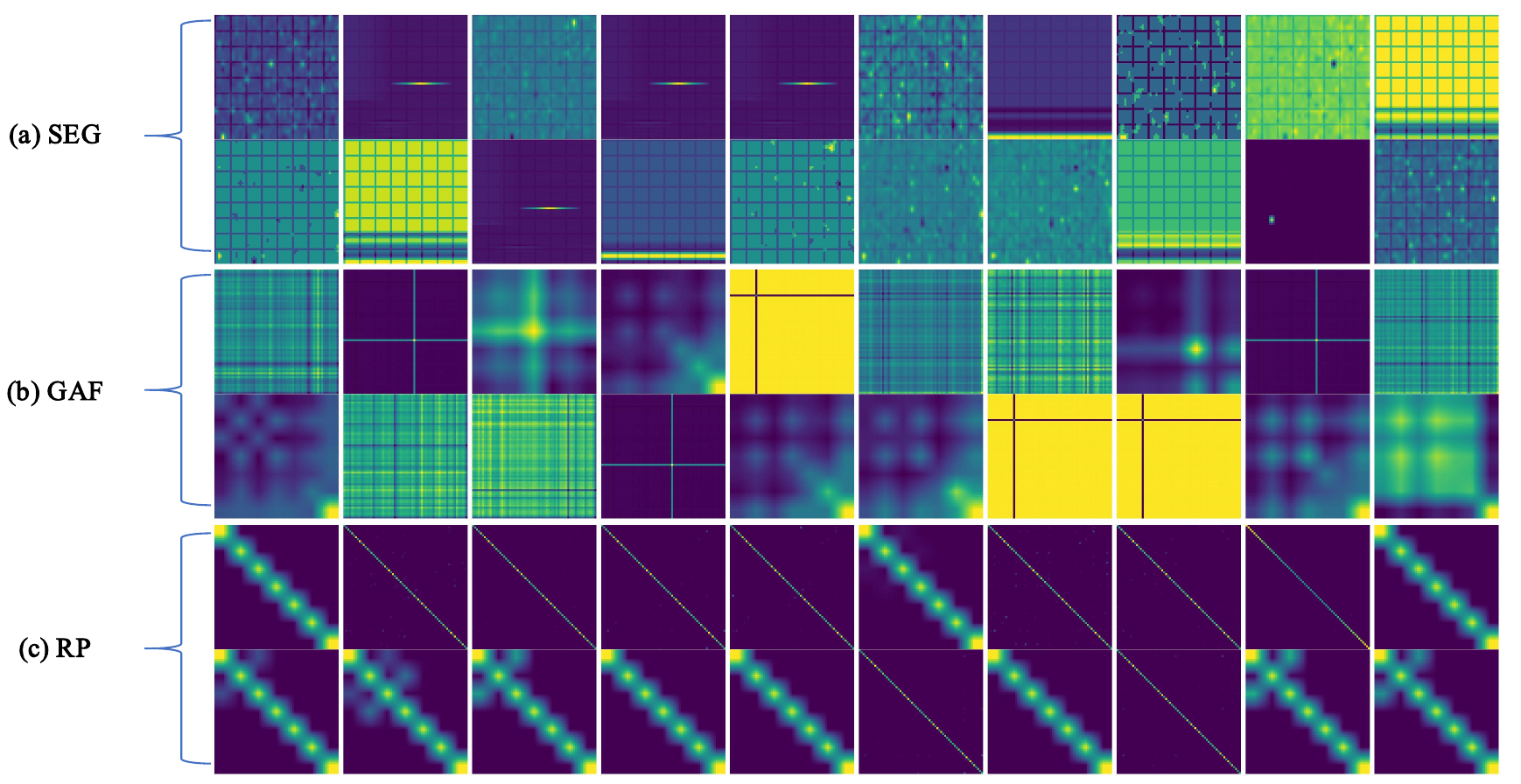}
\caption{Visualization of different time series encoding methods for the vision encoder. We show three approaches: segmentation-based methods (SEG), Gramian Angular Field (GAF), and Recurrence Plot (RP). All methods transform raw time series into image representations with dimensions $\mathbb{R}^{B\times3 \times H\times W}$, where $B$ is the batch size, 3 represents RGB channels, $H$ and $W$ denote the height and width of the generated images.}
\label{fig:visualization_ve}
\end{figure*}

%% file: appendix/C_Algorithm.tex
\newpage
\section{Prerequisites of Latent Diffusion Models}
\label{appx:ldm_algorithm}

\subsection{Autoencoder Framework}
Latent Diffusion Models (LDMs) leverage the autoencoder architecture to facilitate efficient learning in the latent space. An autoencoder comprises two primary components: an encoder and a decoder. The encoder $\mathcal{E}$ compresses high-dimensional input data $x \in \mathbb{R}^D$ into a lower-dimensional latent representation $z \in \mathbb{R}^d$, where $d \ll D$. This compression not only reduces the computational complexity but also captures the essential features of the data. In our implementation, we utilize the pre-trained AutoencoderKL from the \textbf{\textit{stable-diffusion-v1-4}}, which has demonstrated remarkable capabilities in image compression and reconstruction. Mathematically, this process is described as:
\begin{equation}
    z = \mathcal{E}(x)
\end{equation}

\paragraph{Latent Space Scaling}
\label{latent_space_scaling}
In practice, the latent representations produced by the encoder are typically scaled by a factor $s = 0.18215$ to ensure numerical stability and optimal distribution characteristics:
\begin{equation}
    z_{scaled} = s \cdot \mathcal{E}(x)
\end{equation}
This specific scaling factor originates from the VAE design in Stable Diffusion and is derived through empirical optimization. The value is calculated to minimize the KL divergence between the scaled latent distribution and the standard normal distribution:
\begin{equation}
    s^* = \operatorname*{argmin}_{s} \mathbb{E}_{x \sim p_{data}}[D_{KL}(s \cdot \mathcal{E}(x) \| \mathcal{N}(0, 1))]
\end{equation}
where $D_{KL}$ represents the Kullback-Leibler divergence. In our framework, this scaling operation serves multiple critical purposes. It ensures numerical stability during the diffusion process by maintaining consistent value ranges while facilitating better optimization dynamics by bringing the latent distribution closer to the standard normal. This operation also maintains compatibility with the pre-trained weights while allowing for efficient processing of our visual time series representations.

The optimization process involves collecting latent representations $z = \mathcal{E}(x)$ from a large dataset, computing their empirical statistics $\mu_z$ and $\sigma_z^2$, and determining the optimal scaling factor $s$ such that $s\sigma_z \approx 1$ to match the target standard deviation. This process has been extensively validated in the context of both image generation and, in our case, time series visual representations. During decoding, the inverse scaling is applied to restore the original magnitude:
\begin{equation}
    \hat{x} = \mathcal{D}(z_{scaled}/s)
\end{equation}
The autoencoder is trained to minimize the reconstruction loss:
\begin{equation}
    \mathcal{L}_{\text{AE}} = \|\mathcal{D}(\mathcal{E}(x)) - x\|_2^2
\end{equation}
However, in the context of LDMs, the autoencoder enables operations to be performed in the compressed latent space, thereby enhancing efficiency without significant loss of information. In our implementation, we freeze the pret-rained autoencoder parameters in the LDM4TS model during training, focusing the optimization process on diffusion dynamics and temporal feature extraction. This design choice significantly reduces computational overhead while maintaining the benefits of well-learned representations from the compressed latent space.

\subsection{Foundations of Diffusion Models}
Diffusion models define a principled framework for generative modeling through gradual noise addition and removal. In our LDM4TS framework, we adapt this process specifically for time series visual representations while maintaining the fundamental probabilistic structure.

\paragraph{Forward Process}
The forward diffusion process follows a Markov chain that progressively adds Gaussian noise:
\begin{align}
    q(x_t|x_{t-1}) &= \mathcal{N}(x_t; \sqrt{1-\beta_t}x_{t-1}, \beta_t\mathbf{I}) \\
    q(x_t|x_0) &= \mathcal{N}(x_t; \sqrt{\bar{\alpha}_t}x_0, (1-\bar{\alpha}_t)\mathbf{I})
\end{align}
Here, $q(x_t|x_{t-1})$ describes the transition from step $t-1$ to $t$, where $\beta_t$ controls the noise schedule. In our implementation, we adopt a linear noise schedule with carefully tuned parameters $\beta_{start}=0.00085$ and $\beta_{end}=0.012$. The second equation gives the direct relationship between any noisy sample $x_t$ and the original data $x_0$, where $\bar{\alpha}_t = \prod_{s=1}^t (1-\beta_s)$ represents the cumulative product of noise levels.

\paragraph{Reverse Process}
The reverse process learns to gradually denoise data through:
\begin{equation}
    p_\theta(x_{t-1}|x_t) = \mathcal{N}(x_{t-1}; \mu_\theta(x_t,t), \Sigma_\theta(x_t,t))
\end{equation}
where the mean and variance are parameterized as:
\begin{align}
    \mu_\theta(x_t,t) &= \frac{1}{\sqrt{\alpha_t}}(x_t - \frac{\beta_t}{\sqrt{1-\bar{\alpha}_t}}\epsilon_\theta(x_t,t)) \\
    \Sigma_\theta(x_t,t) &= \frac{1-\bar{\alpha}_{t-1}}{1-\bar{\alpha}_t}\beta_t
\end{align}
In our framework, we modify the noise prediction network $\epsilon_\theta$ to accept additional conditioning information, transforming the reverse process into:
\begin{equation}
    p_\theta(x_{t-1}|x_t,c) = \mathcal{N}(x_{t-1}; \mu_\theta(x_t,t,c), \Sigma_\theta(x_t,t))
\end{equation}
where $c$ represents the concatenated frequency domain embeddings and encoded textual descriptions. This modification allows the model to leverage both spectral and semantic information during the denoising process while maintaining the same variance schedule.

\paragraph{Score-based Generation}
The score function represents the gradient of the log-density:
\begin{equation}
    s_\theta(x_t,t) = \nabla_{x_t} \log p_\theta(x_t) = -\frac{\epsilon_\theta(x_t,t)}{\sqrt{1-\bar{\alpha}_t}}
\end{equation}
This formulation enables training through denoising score matching:
\begin{equation}
    \mathcal{L}_{\text{score}} = \mathbb{E}_{t,x_0,\epsilon}[\|\epsilon - \epsilon_\theta(x_t,t)\|_2^2]
\end{equation}

\paragraph{Sampling Methods}
Different sampling strategies offer various trade-offs between generation quality and computational efficiency. In our implementation, we primarily utilize DDIM for its deterministic nature and faster sampling capabilities, though both approaches are supported:
\begin{itemize}
    \item \textbf{DDPM}: Uses the full chain of $T$ steps with stochastic sampling:
    \begin{equation}
        x_{t-1} = \mu_\theta(x_t,t) + \sigma_t\epsilon, \quad \epsilon \sim \mathcal{N}(0,\mathbf{I})
    \end{equation}
    \item \textbf{DDIM}: Enables faster sampling through deterministic trajectories:
    \begin{equation}
        x_{t-1} = \sqrt{\bar{\alpha}_{t-1}}\left(\frac{x_t - \sqrt{1-\bar{\alpha}_t}\epsilon_\theta(x_t,t)}{\sqrt{\bar{\alpha}_t}}\right) + \sqrt{1-\bar{\alpha}_{t-1}}\epsilon_\theta(x_t,t)
    \end{equation}
\end{itemize}

\subsection{U-Net Architecture}
The U-Net architecture serves as the backbone for noise prediction in our framework, combining multi-view processing with skip connections specifically designed for time series visual patterns. Our implementation modifies the standard U-Net structure to better handle temporal dependencies while maintaining spatial coherence.

\paragraph{Encoder-Decoder Structure}
The architecture consists of multiple resolution levels:
\begin{itemize}
    \item \textbf{Downsampling path}: Progressive feature compression
    \begin{equation}
        h_l = \text{ResBlock}(\text{Down}(h_{l-1})), \quad l = 1,\ldots,L
    \end{equation}
    
    \item \textbf{Upsampling path}: Gradual feature reconstruction
    \begin{equation}
        h'_l = \text{ResBlock}(\text{Up}(h'_{l+1})) + h_l, \quad l = L,\ldots,1
    \end{equation}
    
    \item \textbf{Skip connections}: Feature preservation across scales
    \begin{equation}
        h'_l = h'_l + \text{Project}(h_l)
    \end{equation}
\end{itemize}

\paragraph{Feature Extraction}
Each resolution level processes features through a sequence of operations:
\begin{equation}
    F_l = \text{Conv}(\text{GroupNorm}(\text{Attention}(h_l)))
\end{equation}
These operations are augmented with timestep embeddings, which provide temporal information to guide the denoising process:
\begin{equation}
    \gamma_t = \text{MLP}(\text{SinusoidalPos}(t))
\end{equation}
In our implementation, the timestep embedding is projected through a two-layer MLP with SiLU activation, following the design choices in Stable Diffusion for consistency and stability.

\subsection{Attention Mechanisms}
Our model employs attention mechanisms to capture both local and global dependencies in the visual representations:
\paragraph{Self-Attention}
Computes interactions within a feature set through scaled dot-product attention:
\begin{equation}
    \text{Attention}(Q,K,V) = \text{softmax}(\frac{QK^T}{\sqrt{d_k}})V
\end{equation}
where $Q,K,V \in \mathbb{R}^{N \times d_k}$ are query, key, and value matrices.

\paragraph{Cross-Attention}
Enables conditioning through external information:
\begin{equation}
    \text{CrossAttn}(Q,K,V) = \text{softmax}(\frac{QK^T}{\sqrt{d_k}})V
\end{equation}
where $Q$ comes from the main branch and $K,V$ from conditioning information. Multi-head attention extends this:
\begin{equation}
    \text{MultiHead}(Q,K,V) = \text{Concat}(\text{head}_1,\ldots,\text{head}_h)W^O
\end{equation}

\subsection{Conditional Generation}
\label{appx:conditional_generation}
Our framework implements a sophisticated dual-conditioning mechanism that leverages both frequency domain features and semantic descriptions to guide the diffusion process. This multi-modal approach enables robust capture of both temporal patterns and contextual information:

\paragraph{Frequency Conditioning}
To effectively encode the rich spectral information inherent in time series data, we implement a sophisticated frequency domain transformation pipeline. This process begins with the application of a Hann window function, which is crucial for minimizing spectral leakage and enhancing frequency resolution:

\begin{equation}
    w_t = 0.5(1 - \cos(\frac{2\pi t}{L-1}))
\end{equation}
The frequency features are then systematically extracted through a carefully designed three-step process. First, we apply the window function to the input sequence:
\begin{equation}
    X_{win} = X \odot w
\end{equation}

Next, we transform the windowed signal into the frequency domain using the Fast Fourier Transform:
\begin{equation}
    X_{fft} = \text{FFT}(X_{win}) = \sum_{t=0}^{L-1} X_{win}(t) \cdot e^{-2\pi i k t / L}
\end{equation}
Finally, to preserve the complete spectral information, we concatenate the real and imaginary components of the FFT output:
\begin{equation}
    c_{freq} = \text{Concat}[X_{fft_{real}}, X_{fft_{imag}}] \in \mathbb{R}^{B \times (2DL+2)}
\end{equation}
where $L$ denotes the sequence length, $w$ represents the Hann window function, and $\odot$ indicates element-wise multiplication. The terms $X_{fft_{real}}$ and $X_{fft_{imag}}$ correspond to the real and imaginary components of the Fourier transform respectively. This comprehensive encoding strategy enables our model to capture both amplitude and phase information across multiple frequency bands, while maintaining computational efficiency through strategic dimensionality reduction.

\paragraph{Text Conditioning}
To provide semantic guidance for the diffusion process, we automatically generate descriptive prompts by extracting key characteristics from the input time series. The prompt generation function $d_{prompt}(X)$ captures the following statistical properties:
\vspace{-1em}
\begin{itemize}[leftmargin=*, itemsep=0pt]
    \item Statistical measures: minimum, maximum, and median values
    \item Temporal dynamics: overall trend direction and lag patterns
    \item Context information: prediction length and historical window size
    \item Domain knowledge: dataset-specific descriptions
\end{itemize}
\vspace{-1em}
These features are combined into a structured prompt template. A typical generated prompt follows the format:
\begin{quote}
\textit{\textcolor{gray}{"$<|start_prompt|>$Dataset description: \{description\}. Task: forecast the next \{pred\_len\} steps given the previous \{seq\_len\} steps. Input statistics: min value \{min\}, max value \{max\}, median value \{median\}, trend is \{trend\_direction\}, top-5 lags are \{lags\}.$<|<end_prompt>|>$"}}
\end{quote}

The prompts are then processed through a frozen \textit{\textbf{BERT-base-uncased}} model (110M parameters) to extract semantic features. Specifically, each prompt is first tokenized using BERT's WordPiece tokenizer with a maximum sequence length of 77 tokens:
\begin{equation}
    h_{token} = \text{BERT}(d_{prompt}(X)) \in \mathbb{R}^{B \times L_{seq} \times d_{ff}}
\end{equation}
where $L_{seq}$ is the sequence length after tokenization and $d_{ff}=768$ is BERT's hidden dimension. The token-level features are aggregated through mean pooling to obtain a sequence-level representation:
\begin{equation}
    h_{pool} = \frac{1}{L_{seq}}\sum_{i=1}^{L_{seq}} h_{token}[:,i,:] \in \mathbb{R}^{B \times d_{ff}}
\end{equation}
The pooled features are then projected to match the latent dimension through a learnable transformation:
\begin{equation}
    c_{text} = \text{TextEncoder}(X)= \text{TextProj}(h_{pool}) \in \mathbb{R}^{B \times d_{model}}
\end{equation}
where $\text{TextProj}(\cdot)$ consists of a linear layer that projects from $d_{ff}$ to $d_{model}$, followed by layer normalization and ReLU activation to enhance feature expressiveness.

The frequency and text conditions are fused through a cross-modal attention mechanism:
\begin{equation}
    c = \text{CrossAttn}(\text{MLP}([c_{\text{text}}; c_{\text{freq}}])) \in \mathbb{R}^{B \times d_{model}}
\end{equation}
where the MLP first projects the concatenated features to a higher dimension for better feature interaction, and the cross-attention layer enables dynamic feature selection based on the latent representation. This combined conditioning signal guides the diffusion process by injecting both semantic and frequency information into each denoising step through the attention blocks of the U-Net architecture.

%% file: appendix/E_Pseudo.tex
\newpage
\section{Pseudo of LDM4TS framework}
\begin{algorithm}[H]
\caption{Training Process of LDM4TS}
\begin{algorithmic}[1]
\REQUIRE Time series data $\{\mathbf{X}, \mathbf{Y}\}$, where $\mathbf{X} \in \mathbb{R}^{B\times L\times D}$, $\mathbf{Y} \in \mathbb{R}^{B\times pred\_len\times D}$
\ENSURE Trained model parameters $\Theta$

\STATE Initialize VAE encoder $\mathcal{E}$, decoder $\mathcal{D}$, UNet $\mathcal{U}$, \hfill $\triangleright$ Model components
\STATE Initialize diffusion steps $T$, noise schedule $\{\beta_t\}_{t=1}^T$ \hfill $\triangleright$ Diffusion parameters
\STATE $\alpha_t = 1 - \beta_t$, $\bar{\alpha}_t = \prod_{s=1}^t \alpha_s$ \hfill $\triangleright$ Compute coefficients

\WHILE{not converged}
    \STATE Sample mini-batch $\{\mathbf{X}_b, \mathbf{Y}_b\}$ \hfill $\triangleright$ Get training batch
    
    \STATE // Data Preprocessing
    \STATE $means, stdev \gets \text{ComputeStats}(\mathbf{X}_b)$ \hfill $\triangleright$ Calculate statistics
    \STATE $\mathbf{X}_{norm} \gets (\mathbf{X}_b - means) / stdev$ \hfill $\triangleright$ Normalize input
    
    \STATE // Vision Transformation for Time Series
    \FOR{$i \in \{1,\ldots,D\}$}
        \STATE $\mathbf{X}_i \gets \mathbf{X}_{norm}[:,:,i]$ \hfill $\triangleright$ Extract dimension $i$ \;
        \STATE $\mathbf{I}_{seg,i} \gets \text{Segmentation}(\mathbf{X}_i)$ \hfill $\triangleright$ Time series to image \;
        \STATE $\mathbf{I}_{gaf,i} \gets \text{GramianAngularField}(\mathbf{X}_i)$ \hfill $\triangleright$ Polar encoding \;
        \STATE $\mathbf{I}_{rp,i} \gets \text{RecurrencePlot}(\mathbf{X}_i)$ \hfill $\triangleright$ Distance matrix
    \ENDFOR
    
    \STATE $\mathbf{I} \gets \text{Concat}([\mathbf{I}_{seg}, \mathbf{I}_{gaf}, \mathbf{I}_{rp}])$ \hfill $\triangleright$ Combine all views
    
    \STATE // Conditional Controls
    \STATE $\mathbf{f} \gets \text{FreqEncoder}(\text{FFT}(\mathbf{X}_{norm}))$ \hfill $\triangleright$ Extract frequency features
    \STATE $\mathbf{c} \gets \text{TextEncoder}(\text{GeneratePrompt}(\mathbf{X}_b))$ \hfill $\triangleright$ Generate text embedding
    \STATE $\mathbf{cond} \gets \text{FusionLayer}([\mathbf{f}, \mathbf{c}])$ \hfill $\triangleright$ Fuse conditions

    \STATE // Forward Diffusion Process
    \STATE $\epsilon \sim \mathcal{N}(0, \mathbf{I})$ \hfill $\triangleright$ Sample random noise
    \STATE $\mathbf{z}_t \gets \sqrt{\bar{\alpha}_t}\mathbf{z}_0 + \sqrt{1-\bar{\alpha}_t}\epsilon$ \hfill $\triangleright$ Noisy latent
    
    \STATE // Reverse Diffusion Process
    \STATE $\mathbf{z}_0 \gets (\mathbf{z}_t - \sqrt{1-\bar{\alpha}_t}\hat{\epsilon})/\sqrt{\bar{\alpha}_t}$ \hfill $\triangleright$ Denoised latent
    
    
    
    \STATE $\mathbf{z}_0 \gets \mathcal{E}(\mathbf{I})$ \hfill $\triangleright$ Encode to latent space
    \STATE Sample $t \sim \text{Uniform}\{1,...,T\}$ \hfill $\triangleright$ Random timestep
    \STATE $\mathbf{z}_t, \epsilon \gets \text{ForwardDiffusion}(\mathbf{z}_0, t)$ \hfill $\triangleright$ Forward process
    \STATE $\hat{\epsilon} \gets \mathcal{U}(\mathbf{z}_t, t, \mathbf{cond})$ \hfill $\triangleright$ Predict noise
    \STATE $\mathbf{z}_{rec} \gets \text{ReverseDiffusion}(\mathbf{z}_t, \hat{\epsilon}, t)$ \hfill $\triangleright$ Reverse process
    \STATE $\mathbf{I}_{rec} \gets \mathcal{D}(\mathbf{z}_{rec})$ \hfill $\triangleright$ Decode image
    
    \STATE // Feature Extraction and Fusion
    \STATE $\mathbf{h}_v \gets \text{VisionEncoder}(\mathbf{I}_{rec})$ \hfill $\triangleright$ Visual features
    \STATE $\mathbf{h}_t \gets \text{TemporalEncoder}(\text{PatchEmbed}(\mathbf{X}_{norm}))$ \hfill $\triangleright$ Temporal features
    \STATE $\alpha \gets \text{Softmax}(\text{MLP}([\mathbf{h}_v, \mathbf{h}_t]))$ \hfill $\triangleright$ Compute weights
    \STATE $\hat{\mathbf{Y}} \gets \alpha_1\text{VisionHead}(\mathbf{h}_v) + \alpha_2\text{TemporalHead}(\mathbf{h}_t)$ \hfill $\triangleright$ Fuse predictions
    
    \STATE // Optimization
    \STATE $\mathcal{L}_{diff} \gets \|\epsilon - \hat{\epsilon}\|_2^2$ \hfill $\triangleright$ Diffusion loss
    \STATE $\mathcal{L}_{pred} \gets \|\hat{\mathbf{Y}} - \mathbf{Y}_b\|_2^2$ \hfill $\triangleright$ Prediction loss
    \STATE $\mathcal{L} \gets \lambda_1\mathcal{L}_{diff} + \lambda_2\mathcal{L}_{pred}$ \hfill $\triangleright$ Total loss
    \STATE $\Theta \gets \text{Adam}(\Theta, \nabla_{\Theta}\mathcal{L})$ \hfill $\triangleright$ Update parameters
\ENDWHILE

\STATE \textbf{return} $\Theta$ \hfill $\triangleright$ Return trained model
\end{algorithmic}
\end{algorithm}

%% file: appendix/F_Interpretability_of_visual_transformation.tex
\newpage
\section{Analysis of Vision Transformation Methods}
\label{appx:transformation}

Time series analysis faces the fundamental challenge of capturing complex temporal dynamics that manifest simultaneously across multiple scales. While traditional methods excel at specific temporal resolutions, they often struggle to comprehensively capture the full spectrum of patterns ranging from rapid local variations to gradual global trends. This limitation motivates our investigation into vision transformation techniques that can effectively encode rich temporal information into spatial patterns, making them amenable to powerful vision-based processing approaches.

Our framework introduces a systematic approach to time series visualization through three theoretically-grounded transformation methods. Each method targets distinct yet complementary aspects of temporal dynamics, providing a comprehensive representation of the underlying time series structure:

\subsection{Segmentation Representation (SEG)}
The SEG transformation addresses the challenge of preserving local temporal structures while enabling efficient detection of periodic patterns. This method operates by restructuring a time series $x \in \mathbb{R}^L$ into a matrix $M \in \mathbb{R}^{\lceil L/T \rceil \times T}$, where T represents the period length. The transformation process can be formally expressed as:

\begin{equation}
    M_{i,j} = x_{(i-1)T + j}, \quad \text{for } i=1,\dots,\lceil L/T \rceil, j=1,\dots,T
\end{equation}

This segmentation approach offers several theoretical and practical advantages:

\begin{itemize}
    \item \textbf{Local Structure Preservation:} Each row in the matrix represents a complete segment of length T, maintaining the original temporal relationships at the finest granularity
    \item \textbf{Periodic Pattern Detection:} The vertical alignment of segments facilitates the identification of recurring patterns across different time periods
    \item \textbf{Multi-scale Analysis:} By varying the period length T, the transformation can capture patterns at different temporal scales, enabling hierarchical pattern discovery
\end{itemize}

The optimal period length T is determined through an optimization process that maximizes temporal correlation:
\begin{equation}
    T = \arg\max_{k} \sum_{i=1}^{\lceil L/k \rceil} \sum_{j=1}^{k-1} \text{Corr}(M_{i,j}, M_{i,j+1})
\end{equation}

where $\text{Corr}(\cdot,\cdot)$ denotes the correlation between adjacent columns. This optimization ensures optimal alignment of periodic patterns while maintaining temporal fidelity.

\subsection{Gramian Angular Field (GAF)}
The GAF transformation provides a sophisticated approach to encoding temporal relationships through polar coordinate mapping and trigonometric relationships. This method preserves both magnitude and temporal correlation information through a series of carefully designed transformations.

First, the time series is normalized to a bounded interval $[-1,1]$ or $[0,1]$:
\begin{equation}
    \tilde{x}_i = \frac{x_i - \min(x)}{\max(x) - \min(x)}
\end{equation}

The normalized values are then encoded in a polar coordinate system:
\begin{equation}
    \phi = \arccos(\tilde{x}_i), \quad r = \frac{t_i}{N}
\end{equation}

where $t_i$ represents temporal position and $N$ serves as a scaling factor. The final Gramian matrix is constructed through:
\begin{equation}
    G_{i,j} = \cos(\phi_i - \phi_j)
\end{equation}

This transformation offers several key advantages:
\begin{itemize}
    \item \textbf{Scale Invariance:} The polar encoding ensures that the representation is robust to amplitude variations
    \item \textbf{Temporal Correlation Preservation:} The Gramian matrix captures both local and global temporal dependencies
    \item \textbf{Dimensionality Reduction:} The transformation provides a compact representation while preserving essential temporal information
\end{itemize}

\subsection{Recurrence Plot (RP)}
The RP transformation leverages phase space reconstruction to visualize the recurrent behavior in dynamical systems. Based on Taken's embedding theorem, this method first reconstructs the phase space trajectory:

\begin{equation}
    \vec{x}_i = (x_i, x_{i+\tau}, ..., x_{i+(m-1)\tau})
\end{equation}

where $m$ is the embedding dimension and $\tau$ is the time delay. The recurrence matrix is then constructed as:

\begin{equation}
    R_{i,j} = \Theta(\epsilon - \|\vec{x}_i - \vec{x}_j\|)
\end{equation}

where $\Theta$ is the Heaviside function and $\epsilon$ is a threshold distance. This transformation reveals fundamental dynamical properties through several characteristic patterns:

\begin{itemize}
    \item \textbf{Diagonal Lines:} Parallel to the main diagonal, indicating similar evolution of trajectories and revealing deterministic structures
    \item \textbf{Vertical/Horizontal Lines:} Representing periods of state stagnation or laminar phases
    \item \textbf{Complex Patterns:} Non-uniform structures indicating chaos or non-linear dynamics
\end{itemize}

\subsection{Theoretical Integration and Complementarity}
The integration of these three transformations provides a comprehensive framework for time series analysis, offering several theoretical and practical advantages:

\begin{itemize}
    \item \textbf{Multi-scale Pattern Capture:} Each transformation targets different temporal scales - SEG preserves local structures, GAF encodes global correlations, and RP reveals system dynamics
    \item \textbf{Theoretical Complementarity:} The methods maintain theoretical orthogonality while targeting distinct aspects of temporal information
    \item \textbf{Robustness through Diversity:} The combination of different representations provides natural redundancy, mitigating individual limitations
    \item \textbf{Computational Efficiency:} All transformations leverage efficient matrix operations, making them practical for large-scale applications
\end{itemize}

Empirical evidence supports the effectiveness of this multi-view approach, demonstrating superior performance across diverse datasets and prediction horizons compared to single-transformation methods. This success validates our theoretical framework and suggests that comprehensive temporal feature extraction benefits significantly from the synergistic combination of complementary visual representations.

%% file: appendix/G_limitation.tex
\section{Discussion on Applicability and Limitations}
\label{appx:discussion}

\subsection{Theoretical Foundations and Motivations}
The application of Latent Diffusion Models (LDM) to time series forecasting represents a fundamental shift from traditional approaches. This section examines the theoretical foundations and practical motivations behind our framework. Traditional time series forecasting methods often struggle with three key challenges: capturing complex distributions, modeling uncertainty, and handling long-range dependencies. Our LDM-based approach addresses these challenges through its principled probabilistic framework:
\begin{equation}
p(x_{t+1:t+H}|x_{1:t}) = \int p(z)p_\theta(x_{t+1:t+H}|z,x_{1:t})dz
\end{equation}

This formulation offers several advantages. First, it naturally models the full conditional distribution of future values rather than point estimates, enabling robust uncertainty quantification. Second, the iterative denoising process allows the model to capture temporal dependencies at multiple scales. Third, the latent space representation provides an efficient mechanism for learning compressed temporal features.

The integration of vision transformations with LDM is motivated by both theoretical insights and empirical observations. Time series data, when properly transformed into visual representations, exhibits several important properties:

\begin{enumerate}
\item \textbf{Pattern Preservation}: Visual encodings preserve crucial temporal structures including periodicity, trends, and seasonality through spatial arrangements. This is formally expressed as:
\begin{equation}
d(X_1, X_2) \propto d(\mathcal{V}(X_1), \mathcal{V}(X_2))
\end{equation}
where $\mathcal{V}$ represents our vision transformation processing.

\item \textbf{Information Complementarity}: Different visual encodings capture complementary aspects of temporal dynamics:
\begin{equation}
\mathcal{I}(X;Z) \geq \mathcal{I}(X;Z_{visual}) + \mathcal{I}(X;Z_{temporal})
\end{equation}
This property ensures that no critical temporal information is lost during transformation.

\item \textbf{Transfer Learning Potential}: The visual domain enables leveraging powerful pre-trained vision models and their hierarchical feature extraction capabilities.
\end{enumerate}

\subsection{Current Limitations and Future Directions}
The primary limitation of our framework centers on the computational efficiency of the diffusion process. The iterative nature of denoising introduces significant computational overhead, expressed as $T_{\text{total}} = T_{\text{visual}} + t \cdot T_{\text{diffusion}} + T_{\text{projection}}$, which poses challenges for real-time applications and resource-constrained environments. While we have implemented several efficiency-enhancing strategies, such as gradient-free vision transformations and component freezing (including text and vision encoders), the sequential nature of the diffusion process remains a fundamental constraint. The model also exhibits sensitivity to hyperparameter choices in the diffusion schedule, as reflected in the gradient behavior $\sigma(\nabla_\theta \mathcal{L}) \propto \lambda_1\beta_t + \lambda_2\alpha_t$, though this sensitivity is primarily confined to the diffusion component.

Our framework's modular architecture provides clear pathways for future improvements. The design allows for easy replacement and enhancement of individual components, from vision transformation methods (We were able to employ more than the three strategies in the article) to temporal encoder architectures. Future research could focus on developing more efficient training algorithms and incorporating adaptive computation mechanisms while maintaining the model's predictive power. The integration of sophisticated scheduling mechanisms or alternative diffusion formulations could address current computational constraints and improve training stability. Additionally, the framework's extensibility enables adaptation to specific domain requirements and integration of emerging techniques in both visual representation learning and diffusion modeling. These potential improvements, combined with the framework's demonstrated strengths in probabilistic modeling and multi-scale feature capture, suggest promising directions for advancing time series forecasting capabilities.

%% file: appendix/H_Efficiency.tex
\section{Efficiency Analysis}
\label{appx:efficiency}
\begin{table}[h]
\centering
\caption{Computational efficiency comparison on ETTh1 dataset. We report numbers of trainable parameters and inference time (in milliseconds) across different prediction horizons (H).}
\label{tab:efficiency}
\begin{tabular}{r|c|cccc}
\toprule
\multirow{2}{*}{Model} & \multirow{2}{*}{\# Parameters} & \multicolumn{4}{c}{Inference Time (ms)} \\
\cmidrule{3-6}
& & H = 96 & H = 192 & H = 336 & H = 720 \\
\midrule
LDM4TS (Ours) & 5.4M & 76.88 & 80.31 & 193.44 & 192.19 \\
\midrule
TimeGrad & 3.1M & 870.2 & 1854.5 & 3119.7 & 6724.1 \\
ScoreGrad & 4.65M & 3.44 & 4.53 & 4.22 & 7.81 \\ 
CSDI & 10M & 90.4 & 142.8 & 398.9 & 513.1 \\
SSSD & 32M & 418.6 & 645.4 & 1054.2 & 2516.9 \\
\bottomrule
\end{tabular}
\end{table}

We conduct a comprehensive efficiency analysis of LDM4TS, focusing on computational costs and model complexity through extensive experiments on the ETTh1 dataset. Our analysis encompasses both inference time measurements across various prediction horizons and parameter count comparisons with other popular diffusion-based models.

As shown in Table~\ref{tab:efficiency}, LDM4TS demonstrates substantial improvements in inference efficiency over most diffusion-based competitors. For shorter horizons (H=96,192), our model achieves inference times of 76.88ms and 80.31ms respectively, significantly outperforming TimeGrad (870.2ms, 1854.5ms) and SSSD (418.6ms, 645.4ms). Note that for H=720, we observe a slightly faster inference time than H=336, which may be attributed to varying GPU resource availability during our experiments. While ScoreGrad shows faster inference times, our extensive experiments demonstrate that LDM4TS achieves superior forecasting accuracy, offering a better trade-off between efficiency and performance.

The current results indicate that LDM4TS successfully mitigates the computational bottleneck typical of diffusion-based forecasting through our innovative architectural design, including VAE encoding and efficient cross-modal fusion mechanisms. Though traditional models like transformers and linear models maintain faster inference speeds due to their simpler architectures, our ongoing work focuses on further optimization to make diffusion-based forecasting more practical while preserving its superior modeling capabilities.